\documentclass{article}

\usepackage{arxiv}

\usepackage[utf8]{inputenc} % allow utf-8 input
\usepackage[T1]{fontenc}    % use 8-bit T1 fonts
\usepackage{hyperref}       % hyperlinks
\usepackage{url}            % simple URL typesetting
\usepackage{booktabs}       % professional-quality tables
\usepackage{amsfonts}       % blackboard math symbols
\usepackage{nicefrac}       % compact symbols for 1/2, etc.
\usepackage{microtype}      % microtypography
\usepackage{amsmath}        % 提供核心数学功能
\usepackage{amsfonts}       % 提供 \mathbb 等花体字母
\usepackage{amssymb}        % 提供更多数学符号
\usepackage{cleveref}       % smart cross-referencing
\usepackage{lipsum}         % Can be removed after putting your text content
\usepackage{graphicx}
\usepackage{natbib}
\usepackage{multibib}
\newcites{sup}{Supplementary References}
\usepackage{doi}

\title{MetaEarth3D: Unlocking World-scale 3D Generation with Spatially Scalable Generative Modeling}

\date{%
\vspace{-0.8em}
School of Astronautics, Beihang University\\[1.2em]
\small\texttt{Project Page: }\href{https://jinqicao.github.io/metaearth3d}{\texttt{https://jinqicao.github.io/metaearth3d}}\\[1.2em]
}

\newif\ifuniqueAffiliation
\uniqueAffiliationtrue

\ifuniqueAffiliation
\author{%
\vspace{-2.0em}
{Jinqi Cao}\thanks{Equal Contribution}
    \And
    {Zhiping Yu}\footnotemark[1]
    \And
    {Baihong Lin}
    \And
    {Chenyang Liu}
    \And
    {Zhenwei Shi}\thanks{Corresponding author, \textit{shizhenwei@buaa.edu.cn}(Zhenwei Shi), \textit{zhengxiazou@buaa.edu.cn}(Zhengxia Zou)}
    \And
    {Zhengxia Zou}\footnotemark[2]
    \vspace{-0.3em}
}
\fi

\begin{document}
\maketitle

\begin{abstract}
Recent generative AI models have achieved remarkable breakthroughs in language and visual understanding. However, although these models can generate realistic visual content, their spatial scale remains confined to bounded environments, preventing them from capturing how geographic environments evolve across thousands of kilometers or from modeling the spatial structure of the large-scale physical world. This limitation poses a critical challenge for ultra-wide-area spatial intelligence in Earth observation and simulation, revealing a deeper gap in generative AI: progress has relied primarily on scaling model parameters and training data, while overlooking spatial scale as a core dimension of intelligence. Here, motivated by this missing dimension, we investigate spatial scale as a new scaling axis in foundation models and present MetaEarth3D, the first generative foundation model capable of spatially consistent generation at the planetary scale. Taking optical Earth observation simulation as a testbed, MetaEarth3D enables the generation of multi-level, unbounded, and diverse 3D scenes spanning large-scale terrains, medium-scale cities, and fine-grained street blocks. Built upon 10 million globally distributed real-world training images, MetaEarth3D demonstrates both strong visual realism and geospatial statistical realism. Beyond generation, MetaEarth3D serves as a generative data engine for diverse virtual environments in ultra-wide spatial intelligence. We argue that this study may help empower next-generation spatial intelligence for Earth observation.
\end{abstract}

% keywords can be removed

\section{Introduction}
In recent years, large neural network-driven generative foundation models for visual synthesis have attracted extensive attention from both academia and industry (~\cite{1, 1-2, 1-3, 2}). By learning from massive-scale data distributions, such models are able to generate visually realistic images and videos with rich structural and semantic coherence. Alongside the rapid dual scaling in both parameter size and training data volume, their generative capacities have expanded beyond static 2D imagery to simulate increasingly complex real-world scenarios (~\cite{3,4}). Recent breakthroughs have demonstrated highly successful applications across specific domains, including autonomous driving (~\cite{5, 6}), robotics (~\cite{7}), gaming (~\cite{8, 9}), and navigation (~\cite{10}).

\begin{figure}[htbp]
    \centering
    \includegraphics[width=0.95\textwidth]{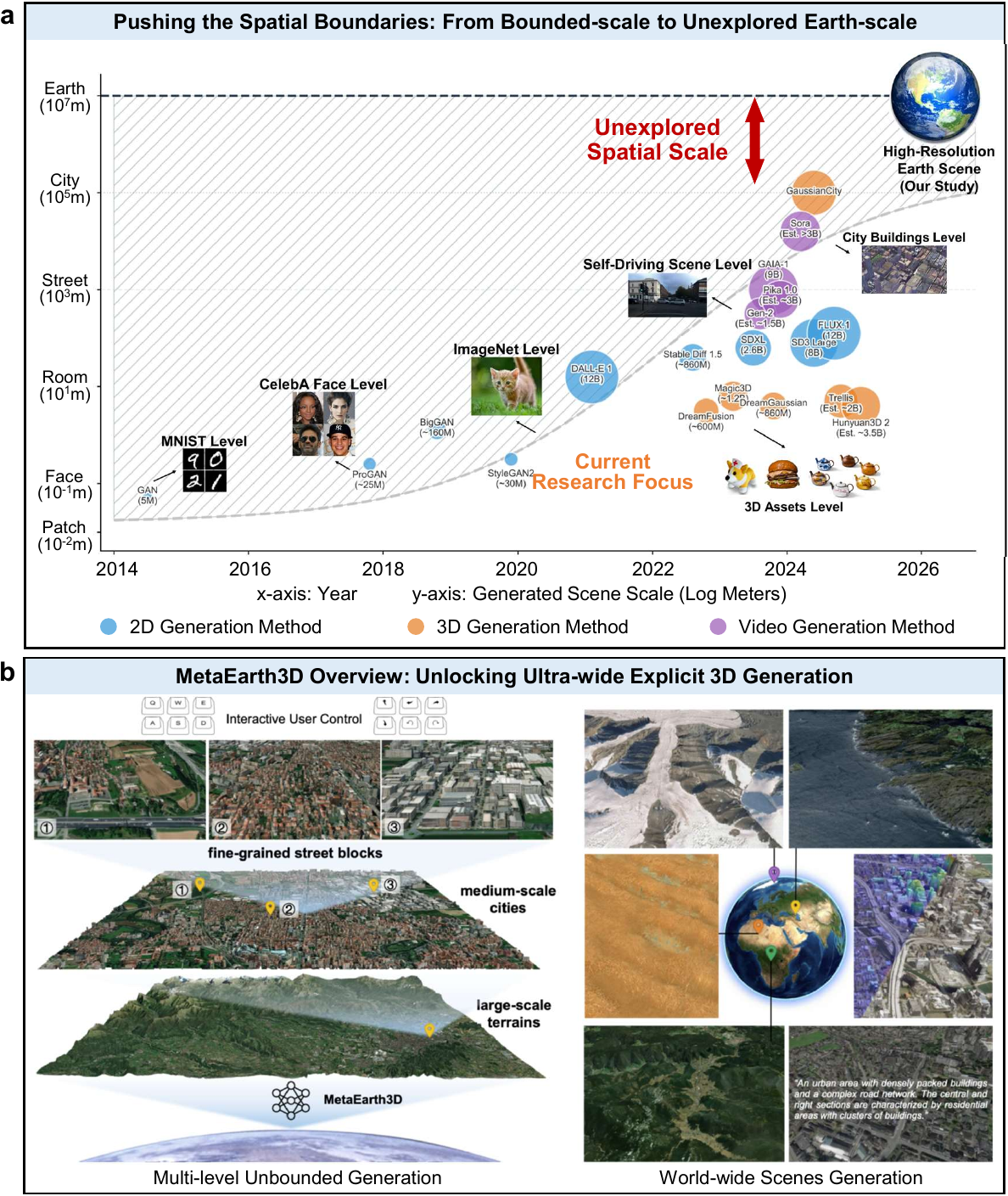}
    \caption{\textbf{Spatial scaling of generative foundation models and the overview of our MetaEarth3D.} \textbf{a,} Chronological evolution of generative models across spatial scales. Circle size and color denote model scale (parameters/data) and generation modality, respectively. Despite the rapid expansion in computational scale, generated environments remain largely confined to object-centric or bounded spatial scales. Detailed references for all plotted methods are provided in Supplementary Table 1. \textbf{b,} Capabilities of MetaEarth3D. The MetaEarth3D extends the spatial boundaries of generative foundation models, unlocking ultra-wide explicit 3D generation. The left panel showcases the powerful capacities in generating multi-level, unbounded scenes, ranging from terrains spanning thousands of square kilometers to medium-scale cites and fine-grained street blocks, supporting continuous observation of user-defined trajectories. The right panel highlights the diversity of worldwide scene generation, ranging from natural landscapes to complex urban structures. MetaEarth3D supports multi-modal conditioning (text or geo-located satellite imagery) and inherently yields self-labeled 3D information, e.g., elevation maps and spatial relationships.}
    \label{fig:fig1}
\end{figure}

Despite the substantial progress achieved in scaling generative foundation models in terms of parameters, data, and modality diversity (~\cite{13, 14, 15, 16}), the spatial scale of these models is restricted to bounded environments, making them challenging in modeling the spatial structure of the large-scale physical world. As shown in Fig. 1a, existing generative models and so-called world models largely remain confined to object-level (~\cite{17}), indoor-level (~\cite{18}), or localized street-view scenes generation (~\cite{19}), with the resolution of synthesized images or videos rarely exceeding 4K (~\cite{20, 21}). More fundamentally, it reveals a deeper gap in the evolution of generative AI: spatial scale, as a core dimension of intelligence, has been largely overlooked. 

As a result, the lack of spatial scalability poses a critical bottleneck for advancing ultra-wide-area spatial intelligence, especially for Earth observation and simulation. Specifically, real-world applications such as advanced air mobility, autonomous aerospace navigation, and disaster management inherently operate across vast geographical expanses. Enabling intelligent perception and decision-making in these domains requires highly realistic virtual environments that span hundreds to thousands of square kilometers (~\cite{22, 23}). Furthermore, tasks like Earth observation digital twins or continuous UAV fly-throughs simulation demand gigapixel-level spatial coverage that can accurately capture continuous transitions from fine-grained urban blocks to expansive natural terrains. Meeting the demand for these applications presents a significant challenge for existing methodologies. Conventional simulation pipelines exhibit distinct limitations: graphics-engine-based simulators (e.g., AirSim (~\cite{24})) provide controllability but lack the textural fidelity and statistical realism of the physical world, while 3D reconstruction strategies (~\cite{25}) are hindered by high costs of data acquisition and restricted scene diversity. Consequently, extending visual generative foundation models to world-scale spatial extents has become a necessary step toward enabling wide-area spatial intelligence.

Recent advancements in generative architectures provide a promising foundation for this pursuit. Historically, visual generative modeling has evolved from early Generative Adversarial Networks (~\cite{26}) to modern diffusion (~\cite{27}) and auto-regressive models (~\cite{28}), which alleviate training instability and mode collapse, enabling the stable training of models with large-scale data and parameters. This expands the generation scope from restricted domains, such as faces or stylized images (~\cite{29}), to general natural scenes at the scale of ImageNet (~\cite{30}). Subsequently, generative modeling has shifted from direct learning in pixel space to compressed latent-space representations (~\cite{31}), where latent diffusion models significantly reduce computational complexity and make high-resolution image and video generation feasible. More recently, diffusion and auto-regressive models (~\cite{32}) built upon large-scale Transformer architectures (~\cite{33}) have further improved generation quality and controllability, while promoting the unification of multimodal and multitask model architectures (~\cite{34}). However, these methods fundamentally operate on rasterized image tokens, whose representational capacity remains bounded to limited spatial domains. When extended to large-area, continuous, and unbounded scenes, the token sequence length grows explosively, which makes training and inference computationally infeasible. Similarly, advanced 3D scene representations such as Neural Radiance Fields (NeRF) (~\cite{35}) and 3D Gaussian Splatting (3DGS) (~\cite{36}) have proven highly effective for object-centric reconstruction, yet encounter severe scalability bottlenecks with prohibitive memory and computational costs when applied to ultra-wide 3D environments. Therefore, identifying efficient and scalable scene representations and generative formulations is essential for unlocking visual generation capabilities at truly ultra-wide spatial scales.

In this paper, we focus on extending the spatial boundaries of generative foundation models. Motivated by optical observation scenarios spanning both on-orbit and low-altitude viewpoints, we investigate how generative models can be scaled from bounded scenes to multi-level and ultra-wide spatial extents. This poses three principal challenges. First, it is fundamentally difficult to construct a unified representation of ultra-wide environment. The Earth’s surface encompasses highly diverse natural and man-made landscapes, including cities, mountains, deserts, forests, and snowfields, each exhibiting distinct geographical characteristics. Even within a single category, such as urban environments, city morphology varies substantially across regions, latitudes, and cultural contexts, and these differences further amplified across multiple observation scales. Compressing such vast and heterogeneous geo-spatial patterns into a single neural model places severe demands on both model capacity and representation efficiency. Second, ultra-wide 3D environments require efficient and spatially consistent scene representations. Existing 3D representations for bounded spaces, such as NeRF, 3D Gaussian Splatting, and point-cloud-based formulations, are intrinsically confined to limited spatial volumes and thus do not naturally extend to kilometer or continent-scale generation. Moreover, large-area continuous environments exhibit complex spatial layouts and long-range structural dependencies, requiring the model to maintain strong spatial coherence across large spatial extents. Third, extending generative modeling to world-scale 3D scenes encounters extreme training complexity. Even 2D generative foundation models already demand substantial computational resources. For example, recent state-of-the-art image generation models such as Stable Diffusion 3 scale to billions of parameters and require large-scale GPU clusters for training. Extending such paradigms to 3D introduces a curse of dimensionality, where volumetric scene complexity grows explosively and renders end-to-end optimization computationally infeasible, thereby calling for more efficient and scalable training formulations.

In response to these challenges, we develop MetaEarth3D, which reformulates ultra-wide 3D scene generation as a progressive transition of probability distributions through coupled scale and dimensional spaces. To improve model capacity and scalability across large spatial ranges, MetaEarth3D adopts a recursive scale-transition generation pipeline within scale space, in which cross-scale 2D images from the same geographical region are generated in a progressive coarse-to-fine manner. A unified model is shared across cascading stages, where each higher-resolution stage is conditioned on the output and spatial resolution of the previous one, enabling efficient parameter reuse and coherent representation of diverse scenes across observation scales. To realize the generation from 2D generated imagery to 3D scenes, we introduce a geometry-texture decoupled dimensional lifting method. A structural geometry generator predicts elevation maps to form a coarse 3D mesh, while a texture generator renders multi-view observations and inpaints missing lateral appearance through multi-view joint generation process. In the texture generator, we further introduce explicit pose-aware conditioning and propose cross-view local attention module to ensure cross-view consistency. By factorizing geometry and appearance, the framework avoids heavy volumetric 3D supervision and learns 3D scene representation directly from 2D imagery, effectively transforming intractable 3D generation task into a set of tractable 2D generative learning tasks. 

Benefiting from the ease of acquiring these 2D imagery, we constructed a 10-million real-world dataset comprising paired images and geographical metadata. The dataset encompasses globally distributed and multi-scale satellite imagery across four spatial resolutions, geo-aligned elevation maps, and multi-view images of urban building textures. This large-scale dataset serves as the foundational bedrock for training MetaEarth3D and validating the effectiveness of the proposed framework. Extensive experiments demonstrate that MetaEarth3D can generate high-fidelity, unbounded, and multi-level 3D environments that span large-scale terrains over thousands of square kilometers, medium-scale cities, and fine-grained street blocks, while supporting continuous observation under interactive user control. Beyond visual realism, MetaEarth3D further achieves geo-distributional statistical realism at the global scale: the high-dimensional semantic feature distributions of the generated satellite imagery and the elevation statistics of the synthesized terrain maps are closely aligned with those of real-world geographical data, indicating that the learned model captures intrinsic geo-spatial regularities. Moreover, the proposed texture generator endows MetaEarth3D with strong spatial and multi-view consistency across large-area scenes, such that generated 3D scenes maintain highly coherent textures across viewpoints. Our method achieves memory-efficient and rapid scene generation, allowing large-scale scenes to be generated on consumer-grade GPUs. For instance, using a single NVIDIA RTX 4090 (24 GB), MetaEarth3D can generate a district-scale scene (17 km²) within only 2 hours.

Unlike video-based generative models that generates scenes of implicit representations, MetaEarth3D generates explicit 3D meshes enriched with self-labeled native 3D annotations, making it readily amenable to deployment as a generative data engine and integration with simulation environments for Earth observation and intelligent aerospace platforms. To quantitatively demonstrate this practical value, we choose ultra-wide visual perception and reasoning as representative downstream tasks, including spatial, morphology, counting, geometry and captioning, spanning five distinct facets of 3D scene cognition. Research in this domain remains limited in real-world scenarios, primarily due to the scarcity of high-quality 3D ground truth and the high cost of data collection and annotations. MetaEarth3D bridges this gap and unlocks the potential for such research by (i) generating diverse natural and man-made environments as explicit meshes, (ii) rendering calibrated UAV-view RGB observations along controlled trajectories, and (iii) exposing native mesh-derived 3D supervision (e.g., height statistics and structured 3D relations) that cannot be reliably obtained from real imagery alone. Utilizing MetaEarth3D as a generative data engine, we constructed a wide-area spatial reasoning dataset comprising 60 scenes and 7690 samples based on synthetic paired 2D observations and verifiable 3D supervision. We fine-tuned open-source 2D vision-language models (~\cite{37, 38, 39}) with the synthesized dataset and test the fine-tuned models on real-world UAV scenes, where we observe an average improvement of +22.85\% across five typical geospatial understanding tasks, effectively extending the visual perception capabilities of these models into real-world 3D scenarios. Building on these foundations, we believe MetaEarth3D provides a scalable and practically deployable pathway toward advancing ultra-wide spatial intelligence in the long term.

\section{Methods}

\subsection{Progressive Probabilistic Framework for Ultra-wide 3D Scene Generation}

Fundamentally, the generation of a large-scale 3D scene $\mathcal{X}$ is equivalent to modeling and sampling from its underlying probability distribution $p(\mathcal{X})$. However, directly modeling this distribution is computationally intractable, particularly for ultra-wide scenes exhibiting diverse distributional characteristics, due to the extreme dimensionality and the complex correlations between components. To address this challenge, we formulate the generation process as a sequential transition through two coupled physical spaces: the scale space, which models the coarse-to-fine generation process of multi-resolution remote sensing imagery, and the dimensional space, which then progressively lifts the scene dimensionality from 2D to 3D.

\textbf{Progressive Lifting at Dimensional Space.} To avoid direct modeling of the full 3D scene, we decouple $\mathcal{X}$ into three distinct components: the orthographic imagery $x_o^{(k)}$, the elevation geometry $x_h$ and the lateral appearance $x_l$, denoted as $\mathcal{X} = \{x_o^{(k)}, x_h, x_l\}$. Specifically, $x_o^{(k)}$ refers to satellite remote sensing imagery at spatial resolution $s^{(k)}$, representing the visual content of the 3D scene observed from a bird's eye view at a specific observation scale. $x_h$ represents the height values of the terrain and ground objects, and $x_l$ represents the visual textures of the vertical surfaces. Through this decoupling, we can factorize the intractable high-dimensional distribution into a chain of conditional probabilities:
\begin{equation}
    p(x_o^{(k)}, x_h, x_l) = p(x_o^{(k)}) \cdot p(x_h | x_o^{(k)}) \cdot p(x_l | x_o^{(k)}, x_h)
\end{equation}

\begin{figure}[htbp]
    \centering
    \includegraphics[width=\linewidth]{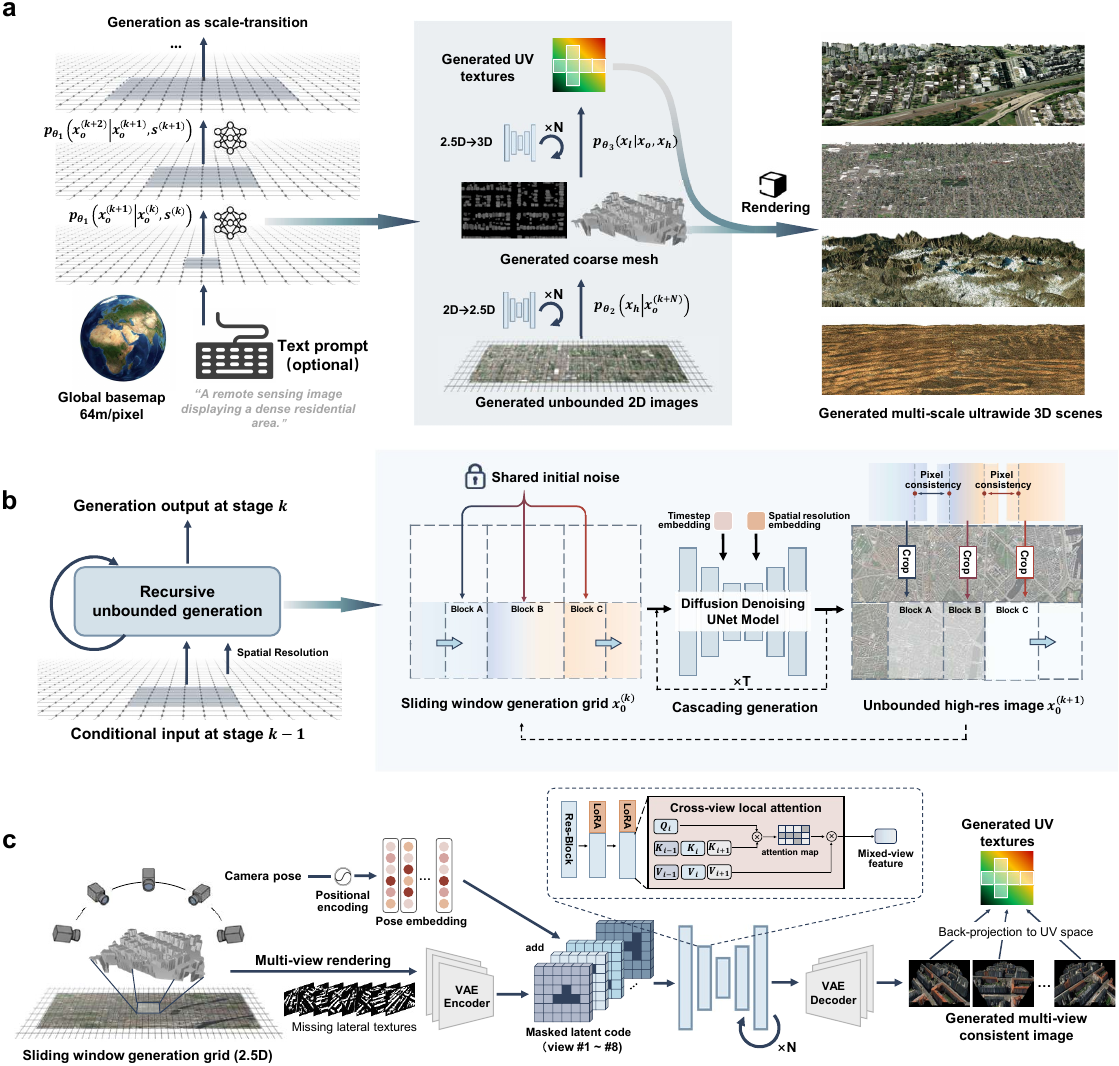}
    \caption{\textbf{MetaEarth3D generative framework and model architecture.} \textbf{a,} The overall progressive probabilistic generative framework for ultra-wide 3D scene generation. This pipeline illustrates the transformation of real imagery or text prompts into a generated large-scale 3D mesh. The process is divided into scale space transition and dimensional space lifting. \textbf{b,} Recursive multi-scale satellite imagery generation module. This module details the self-casaded mechanism for ultra-wide unbounded image generation. \textbf{c,} Multi-view lateral appearance inpainting. By embedding camera pose information and incorporating cross-view local attention into the model, the model generates multi-view consistent images, which are back-projected onto the UV map to refine the final 3D mesh.}
    \label{fig:fig5}
\end{figure}

Accordingly, we construct the MetaEarth3D, a generative foundation model to model this chain of conditional probabilities. By modeling the conditional dependencies and distributions among visual semantics, spatial geometry, and texture, we realize a progressive dimensional lifting process from 2D images to 3D scenes. Fig.~\ref{fig:fig5}a illustrates the model architecture and generation pipeline of MetaEarth3D. Specifically, the generation process consists of three stages. Given MetaEarth3D with parameters $\theta = \{\theta_1, \theta_2, \theta_3\}$, where $\theta_1$, $\theta_2$ and $\theta_3$ denote the network parameters corresponding to the three generation stages respectively. First, the model generates satellite imagery $x_o^{(k)} \sim p_{\theta_1} (x_o^{(k)})$ at a specific spatial resolution $s^{(k)}$, establishing the semantic layout on the 2D plane for the target observation scale. Subsequently, conditioned on the generated $x_o^{(k)}$, the model infers the height map $x_h \sim p_{\theta_2} (x_h | x_o^{(k)})$, lifting the 2D imagery to form a 2.5D structure. Finally, jointly conditioned on the satellite imagery $x_o^{(k)}$ and the elevation map $x_h$, the model generates the lateral appearance $x_l \sim p_{\theta_3} (x_l | x_o^{(k)}, x_h)$ to inpaint the missing vertical textures. This process can ultimately generate a realistic 3D scene from a sample satellite imagery represented as an explicit mesh.

\textbf{Coarse-to-Fine Resolution Refinement at Scale Space.} Serving as the foundational semantic anchor for the subsequent dimensional lifting, the orthographic imagery $x_o^{(k)}$ determines the structural plausibility and visual fidelity of the final 3D scene. To efficiently represent diverse global scene features and ensure multi-level consistency in base satellite image generation, we draw inspiration from the self-similarity of geospatial terrain features within the scale space. Consequently, we formulate the generation of the target imagery as a sequential distribution transition within the scale space. Mathematically, we model this process as a Markov chain transition evolving from low-spatial-resolution to high-spatial-resolution images. Let $\mathcal{X}_o = \{x_o^{(k)}, x_o^{(k+1)}, \dots, x_o^{(K)}\}$ denote a sequence of orthographic images with spatial resolutions increasing by a factor of $N$. Specifically, if $x_o^{(i)}$ possesses a spatial resolution of $s^{(i)}$ (m/pixel) with $H \times W$ pixels, then $x_o^{(i+1)}$ has a spatial resolution of $s^{(i+1)} = N s^{(i)}$ with $N H \times N W$ pixels. Since high-resolution imagery subsumes the information of lower resolutions, it is reasonable to assume that the generation of the image at level $i + 1$ depends solely on the image at level $i$. By introducing this Markov assumption, the joint distribution $p_{\theta_1}(\mathcal{X}_o)$ is factorized as:
\begin{equation}
    p_{\theta_1} (x_o^{(k)}, x_o^{(k+1)}, \dots, x_o^{(K)}) = p(x_o^{(k)}) \cdot \prod_{i=k}^{K} p_{\theta_1} (x_o^{(i+1)} | x_o^{(i)}, s^{(i+1)})
\end{equation}

This factorization decomposes the multi-scale image generation into a self-cascading coarse-to-fine resolution refinement process operating in the scale space. Fundamentally, this design leverages the high correlation between adjacent scales to maximize representation efficiency. Since the visual content evolves gradually across resolutions, a single network can effectively learn the differential increments at each step. This allows for parameter reuse, enabling the model to characterize complex multi-scale features with a unified set of parameters rather than independent models. Specifically, the process is initialized with a starting image $x_o^{(k)}$. This initial anchor can either be sampled from the real-world distribution (i.e., existing remote sensing imagery) or generated via a generative model, such as a text-to-image diffusion model. The term $\prod_{i=k}^{K} p_{\theta_1} (x_o^{(i+1)} | x_o^{(i)}, s^{(i+1)})$ formalizes the cascaded generation using a single shared neural network. The model is conditioned on two inputs: the previous state $x_o^{(i)}$ and the target resolution $s^{(i+1)}$. The former ensures multi-level consistency, allowing the high-resolution generation adheres to the structural and semantic constraints established by the low-resolution input. The latter equips the model with scale-awareness, enabling the shared neural network to adaptively generate features appropriate for the current resolution.

In the following sections, we detail the specific module architectures instantiating these probabilistic processes, sequentially introducing the scale-space generation module ($\theta_1$) and the joint dimensional lifting modules ($\theta_2, \theta_3$).

\subsection{Recursive Multi-scale Satellite Imagery Generation Module}

\textbf{Resolution-guided Recursive Diffusion Network.} To effectively model the Markov chain transition $p_{\theta_1} (x_o^{(i+1)} | x_o^{(i)}, s^{(i+1)}), i = k, \dots, K$ within the scale space, we employ a resolution-guided recursive diffusion network. Fig.~\ref{fig:fig5}b illustrates the cascading generation workflow for multi-scale imagery within our MetaEarth3D framework. We adopt the denoising diffusion probabilistic model (DDPM)~\cite{27} as the generative backbone, which can be formulated through two inverse processes: the forward noise injection and the reverse denoising process. The forward noise injection process gradually corrupts the clean data $x_0$ by injecting Gaussian noise over $T$ discrete timesteps. At any step $t$, the transition is defined as: $q(x_t^{(i+1)} | x_{t-1}^{(i+1)}) = \mathcal{N}(x_t^{(i+1)}; \sqrt{1-\beta_t} x_{t-1}^{(i+1)}, \beta_t \mathbf{I})$, where $x_t^{(i+1)}$ and $x_{t-1}^{(i+1)}$ denote the latent variables of the target image $x_o^{(i+1)} \in \mathbb{R}^{H \times W \times 3}$ at timesteps $t$ and $t-1$ respectively, and $\beta_t$ represents the variance schedule. As $T \to \infty$, the latent variable $x_T$ converges to a standard isotropic Gaussian $\mathcal{N}(0, \mathbf{I})$. The reverse denoising process (i.e., the generative process) is defined as a learnable reverse trajectory that recovers $x_0$ from standard gaussian noise $\epsilon \sim \mathcal{N}(0, \mathbf{I})$. Following the standard DDPM, we approximate the intractable posterior using a conditional Gaussian transition:
\begin{equation}
    p_{\theta_1} (x_{t-1}^{(i+1)} | x_t^{(i+1)}, c^{(i+1)}) = \mathcal{N} \left( x_{t-1}^{(i+1)}; \mu_{\theta_1} (x_t^{(i+1)}, t, c^{(i+1)}), \sigma_t^2 \mathbf{I} \right)
\end{equation}
where the variance $\sigma_t^2$ is set to fixed time-dependent constants, while the $\mu_{\theta_1}$ is derived from the noise predicted by a denoising neural network. The condition set $c^{(i+1)} = \{x_o^{(i)}, s^{(i+1)}\}$ consists of the input low-resolution image and the target spatial resolution. To implement this learnable conditional reverse transition, we adopt a high-capacity UNet-like neural network (approximately 1 billion parameters) from our previous work as the denoising backbone. The detailed architectural configuration is provided in the Supplementary Materials. We further design two specific encoding branches to integrate the semantic and scale constraints into the next-scale image generation.

Specifically, a dedicated image encoder $E_{lr}$ (stacked convolutional layers) extracts semantic features from $x_o^{(i)}$. These features are spatially aligned with $x_t^{(i+1)}$ via upsampling layers $F_{up}$ and subsequently fused with the noisy latent $x_t^{(i+1)}$ through channel-wise concatenation: $\tilde{x}_t^{(i+1)} = \text{Concat} \left[ x_t^{(i+1)}, F_{up} \left( E_{lr}(x_o^{(i)}) \right) \right]$, serving as the joint input to the denoising network. To empower the denoising network with scale awareness, we treat the target spatial resolution $s^{(i+1)}$ as a continuous variable and project it into a high-dimensional embedding space using sinusoidal positional encodings. Similar to the timestep embedding in standard DDPM, this resolution embedding interacts with the network features via adaptive group normalization, informing the model of the current scale level and guiding the generation of appropriate textural frequencies.

\textbf{Ultra-wide Unbounded Image Generation Method.} The cascaded generation process described above represents an idealized formulation, implicitly assuming that the model can process inputs and outputs of arbitrary dimensions without memory constraints. However, in practice, directly generating ultra-wide, high-spatial-resolution scenes faces severe challenges: the prohibitive GPU memory footprint makes training and inference computationally intractable, and the fixed-size architecture inherently lacks the flexibility to handle unbounded spatial inputs. To overcome these limitations and achieve arbitrary-size image generation, we adopt an unbounded generation method. To strictly control the computational overhead, we spatially decompose the ultra-wide generation task into a grid of manageable, fixed-size generation sub-tasks. Specifically, we constrain the input image dimensions to a fixed scale (e.g., $64 \times 64$), partitioning the large-scale condition into local patches for independent generation, which are subsequently reassembled into the complete image. However, due to the inherent indeterminacy of diffusion processes, naive tiling leads to discontinuities in global semantics and texture, resulting in visible boundary artifacts. To eliminate these inconsistencies, we reformulate the reverse denoising process as a deterministic probability flow ODE (e.g., DDIM sampling strategy (~\cite{50})) rather than a stochastic Markov chain. Under this formulation, the generated image is strictly controlled by the initial noise $x_T \sim \mathcal{N}(0, \mathbf{I})$ and the guidance $c$. Based on this property, we traverse the scene using sliding-windows with 50\% overlap and enforce that identical initial Gaussian noise is used for the overlapping regions of adjacent windows. Due to the deterministic nature of the ODE solver, identical initial latents within these overlaps yield pixel-wise consistent content. This mathematically guarantees that locally independent inference processes seamlessly merge into a visually consistent and continuous unbounded image.

\subsection{Geometry-Texture Decoupled Dimensional Lifting Module}

Having obtained high-fidelity orthographic imagery $x_o$ at the target observation scale, MetaEarth3D further lifts the generated $x_o$ into an explicitly 3D scene in dimensional space. To bridge this dimensionality gap, we propose a geometry-texture decoupled dimensional lifting module (as shown in Fig.~\ref{fig:fig5}c) and decompose the intractable 3D generation task into two sequential conditional generation sub-tasks: generative height inference ($\theta_2$), which predicts the vertical geometry $x_h$, and lateral appearance inpainting ($\theta_3$), which generates the missing textures $x_l$ for vertical surfaces.

\textbf{Generative Height Inference.} The first stage of dimensional lifting aims to infer the underlying terrain and object geometry, instantiated as the conditional distribution $p_{\theta_2}(x_h | x_o)$. While height estimation from a single view is theoretically ill-posed, satellite imagery is rich in monocular cues, such as shadows and perspective distortions, that implicitly constrain the solution space, making probabilistic inference feasible. To model this distribution, we reformulate height estimation as a prompt-conditioned image-to-image translation task (~\cite{51}), map visual features and monocular cues in $x_o$ directly to geometric structures in $x_h$ with text prompt "predict the heights of prominent features". Specifically, $x_o$ is encoded into the latent space via a pretrained VAE (~\cite{31}) encoder and concatenated with the noisy latents of the target elevation map. The text prompt is introduced to explicitly define the generation task, making the diffusion model focus on geometric prediction and suppress unrelated image synthesis behaviors. The model is trained using the standard noise prediction objective, i.e., minimizing the $L_2$ distance between sampled Gaussian noise $\epsilon$ and the noise predicted by the network $\epsilon_{\theta_2}$:
\begin{equation}
    \mathcal{L}_{\theta_2} = \mathbb{E}_{z_0, x_o, \tau, t, \epsilon \sim \mathcal{N}(0, \mathbf{I})} \left[ \left\| \epsilon - \epsilon_{\theta_2} (z_t, t, \tau, x_o) \right\|_2^2 \right]
\end{equation}
where $z_t$ represents the noisy latent of the height map $x_h$ at timestep $t$, and the $\tau$ represents text instruction. To adjust orthographic image inputs $x_o$ of varying pixel sizes and generate correspondingly sized height maps $x_h$, we extend the unbounded generation method into the latent space. Specifically, we apply overlapping sliding-window cropping to the input $x_o$ in the pixel space. Premised on the property that the VAE encoding and decoding process preserves spatial structural consistency, meaning the relative positional relationships of the overlapping regions remain invariant, we enforce that the initial noise sampled in the latent space is identical for these spatially corresponding overlapping areas. This strategy effectively achieves the generation of unbounded and continuous height maps.

\textbf{Multi-view Lateral Appearance Inpainting.} The predicted height map $x_h$ and the orthographic image $x_o$ can be combined via geometric projection to synthesize a coarse-grained 3D mesh. However, in this coarse mesh, vertical geometric structures (e.g., building facades) lack visual content and texture; these regions correspond to the blind spots $x_l$ under the satellite view. To complete the 3D scene, we aim to generate these missing textures $x_l$, modeled as the conditional distribution $p_{\theta_3} (x_l | x_o, x_h)$. Unlike independent image inpainting or restoration, the generation of $x_l$ requires strict adherence to multi-view spatial consistency. That is, as the observational viewpoint continuously changes, the lateral textures must remain visually stable and continuous. To model complex spatial relationships between views, we propose a multi-view texture joint generation latent diffusion module. Our model shares architectural similarities with AnimateDiff (~\cite{52}), repurposing its 3D attention modules to model cross-view spatial correlations. Specifically, we first partition the large-scale coarse 3D mesh into multiple memory-feasible block regions. Then, we render a set of multi-view images $\mathcal{V} = \{v^{(1)}, v^{(2)}, \dots, v^{(N)}\}$ from $N$ equidistant viewpoints surrounding the scene trajectory (e.g., 8 viewpoints spaced by $45^\circ$). For each viewpoint $v^{(i)} \in \mathbb{R}^{H \times W \times 3}$, vertical regions lacking texture are identified via surface normal analysis and marked with a binary mask $M^{(i)} \in \mathbb{R}^{H \times W \times 1}$. At each diffusion timestep $t$, the model inputs can be expressed as: $\mathcal{U}_t = \left[ u_t^{(1)}, u_t^{(2)}, \dots, u_t^{(N)} \right] \in \mathbb{R}^{N \times h \times w \times (2C+1)}$, where $u_t^{(i)} = \text{Concat} \left[ z_t^{(i)}, \mathcal{E}(v^{(i)}), m^{(i)} \right]$, consisting of three components: the noisy latent $z_t^{(i)} \in \mathbb{R}^{h \times w \times C}$ for the $i$-th viewpoint, the VAE-encoded features $\mathcal{E}(v^{(i)}) \in \mathbb{R}^{h \times w \times C}$ of the viewpoint $v^{(i)}$, and the downsampled binary mask $m^{(i)} \in \mathbb{R}^{h \times w \times 1}$ for the $i$-th viewpoint. To explicitly endow the model with the ability to perceive viewpoint variations and 3D space, we inject camera pose information into the model. Since the multi-view images are rendered along a fixed circular trajectory with equidistant intervals, the relative geometric relationships between adjacent viewpoints are fixed. This allows us to efficiently represent the camera pose using the viewpoint index instead of complex transformation matrices. Specifically, we map the index $i$ of each view to a learnable viewpoint embedding $e^{(i)} \in \mathbb{R}^C$, and the embedding is then added to the initial convolutional feature maps of the network:
\begin{equation}
    F_{in}^{(i)} = \text{Conv}_{in} (u^{(i)}) + e^{(i)}
\end{equation}
where $\text{Conv}_{in}$ is the input convolution layer of the diffusion backbone. By injecting $e^{(i)}$ at the input stage, we break the permutation invariance of the sequence, explicitly grounding each view to its physical orientation before the multi-view interaction begins. To further explicitly ensure spatial consistency across multi-view images, we propose a novel cross-view local attention module. We observe that adjacent viewpoints share significant visual overlap, which serves as the physical basis for consistency. Standard global attention is computationally expensive and introduces noise from non-overlapping distant views, while independent self-attention fails to maintain continuity. Therefore, we redesign the attention layers to focus on local spatial neighborhoods. Formally, let $q_i \in \mathbb{R}^d$ denote the query token for the $i$-th view. The attention operation is restricted to interact only with the keys $k$ and values $v$ from itself and its neighbors, i.e. $N(i) = \{i-1, i, i+1\}$ (using circular indexing for the $360^\circ$ loop):
\begin{equation}
    \text{Attention}(q_i, \mathcal{K}, \mathcal{V}) = \text{Softmax} \left( \frac{q_i (k_{N(i)})^{\mathsf{T}}}{\sqrt{d}} \right) v_{N(i)}
\end{equation}
where $k_{N(i)}$ and $v_{N(i)}$ represent the concatenated keys and values from the current and adjacent views. By allowing the query $q_i$ to "see" the visual content of its neighbors, the model effectively models shared texture patterns within overlapping fields of view, resulting in the multi-view continuous and visually consistent texture completion. Based on the aforementioned method, MetaEarth3D generates a set of images $\mathcal{J} = \{I^{(1)}, I^{(2)}, \dots, I^{(N)}\}$ characterized by multi-view texture continuity and visual consistency.

\textbf{Texture Back Projection via Multiview Geometric Constraints.} To map the inpainted 2D multi-view images $\mathcal{J}$ back onto the 3D mesh surface, we propose an inverse projection and texture baking framework based on normal vector selection. This method establishes a precise and high-quality mapping between 3D surface points and 2D generated images via inverse projection, preserving the existing texture of the coarse mesh while utilizing the generated multi-view images to complete the missing lateral appearance $x_l$. The coarse-grained 3D mesh is formally defined as $\mathcal{M} = (\mathcal{P}, \mathcal{F})$, where $\mathcal{P} \subset \mathbb{R}^3$ represents the set of discrete vertex coordinates defining the scene geometry, and $\mathcal{F}$ represents the set of triangular faces. To isolate the regions requiring inpainting, we decompose the faces $\mathcal{F}$ into a set of vertical surfaces $\mathcal{F}_{vert}$ and horizontal surfaces $\mathcal{F}_{hori}$. Specifically, for any face $f_i \in \mathcal{F}$, we calculate the dot product between its unit normal vector $\mathbf{n}_i$ and the vertical axis $z_{up}$ of the world system. By introducing a threshold $\tau$, we partition the faces into vertical and horizontal categories as follows:
\begin{equation}
    f_i \in 
    \begin{cases} 
        \mathcal{F}_{vert}, & \text{if } |\mathbf{n}_i \cdot z_{up}| < \tau \\ 
        \mathcal{F}_{hori}, & \text{otherwise} 
    \end{cases}
\end{equation}
where $\mathcal{F}_{hori}$ retains the original orthographic texture, while $\mathcal{F}_{vert}$ is marked as the region to be inpainted. We perform texture parameterization and atlas packing specifically for $\mathcal{F}_{vert}$ to allocate independent coordinate regions in the 2D texture space. To transfer multi-view image information to the texture atlas without introducing holes, we employ an inverse baking strategy. For each discrete texel $T$ in the texture atlas, we first retrieve its corresponding vertex coordinates $P_j \in \mathcal{P}$. Subsequently, based on the pinhole camera model, we map $P_j$ to the pixel coordinates $x_{j,k} = (x, y, 1)^{\mathsf{T}}$ in the $k$-th view image $I^{(k)}$, computed as:
\begin{equation}
    s_{proj} \cdot x_{j,k} = K [R_k | t_k] \cdot P_j
\end{equation}
where $K, R_k$, and $t_k$ denote the camera intrinsics, rotation, and translation matrices for the $k$-th view $I^{(k)}$, and $s_{proj}$ is the projective scale factor (i.e., depth). To address occlusion and view selection issues, we introduce a visibility-aware selection strategy. For the surface point $P_j$, we perform visibility checks using Z-buffering to identify the set of visible view indices $\mathcal{K}_{vis}$. With this set, we select the optimal view index $k^*$ that minimizes the viewing angle, which is equivalent to maximizing the dot product between the surface normal $\mathbf{n}_j$ and the normalized viewing direction $d_{view}^{(k)}$:
\begin{equation}
    k^* = \operatorname*{argmax}_{k \in \mathcal{K}_{vis}} \left( \mathbf{n}_j \cdot d_{view}^{(k)} \right)
\end{equation}
The texture is then sampled from the optimal image $I^{(k^*)}$ at coordinates $(x, y)$ via bilinear interpolation and baked into the corresponding texel $T$. This ensures that each surface patch is textured by the most perpendicular observation, achieving high-fidelity back-projection.

\subsection{Evaluation metrics for MetaEarth3D}

\textbf{Evaluation metrics for generative fidelity.} We employ the Fr{\'e}chet Inception Distance (FID) (~\cite{53}) to measure the fidelity of generated imagery:
\begin{equation}
    FID = \left\| \mu_r - \mu_g \right\|^2 + \operatorname{Tr} \left( \Sigma_r + \Sigma_g - 2 (\Sigma_r \Sigma_g)^{1/2} \right)
\end{equation}
where $(\mu_r, \Sigma_r)$ and $(\mu_g, \Sigma_g)$ represent the mean and covariance of deep features extracted from a pre-trained neural network (e.g., InceptionV3). $\operatorname{Tr}(\cdot)$ denotes the trace of a matrix. A lower FID indicates higher generation quality.

\textbf{Evaluation metrics for spatial and multi-view consistency.} We quantify geometric continuity using the Mean Seam Gradient (MSG), which calculates the average absolute pixel difference across stitching boundaries:
\begin{equation}
    MSG = \frac{1}{M} \sum_{i=1}^{M} \left| \nabla_{boundary} x_i \right|
\end{equation}
where $|\nabla_{boundary} x_i|$ denotes the absolute gradient magnitude of the pixel $x_i$ at the image stitching boundary and $M$ represents the total number of pixels. A lower MSG score indicates that the generated height map possesses better spatial consistency. For lateral appearance, it is considered consistent if the texture $I_{src}$ generated from a source view can be reprojected to a target view via the generated geometry without significant deviation from the directly generated target view $I_{tgt}$. We quantify this using the Peak Signal-to-Noise Ratio (PSNR). First, we calculate the Mean Squared Error (MSE) between the reprojected image $I_{src}$ and the target image $I_{tgt}$ over the $N$ overlapping pixels:
\begin{equation}
    MSE = \frac{1}{N} \sum_{i=1}^{N} \left\| I_{tgt}^{(i)} - I_{src}^{(i)} \right\|^2
\end{equation}
Based on the MSE, the PSNR is rigorously defined as:
\begin{equation}
    PSNR = 10 \cdot \log_{10} \left( \frac{MAX_I^2}{MSE} \right)
\end{equation}
where $MAX_I$ represents the maximum possible pixel intensity. A higher PSNR value indicates minimal geometric distortion and superior multi-view consistency.

To capture perceptual structural similarity beyond pixel-level errors, we further employ the Structural Similarity Index Measure (SSIM), which models image degradation as perceived changes in structural information. It is calculated as:
\begin{equation}
    SSIM(x, y) = \frac{(2\mu_x\mu_y + C_1)(2\sigma_{xy} + C_2)}{(\mu_x^2 + \mu_y^2 + C_1)(\sigma_x^2 + \sigma_y^2 + C_2)}
\end{equation}
where $\mu_x, \mu_y$ are the average intensities, $\sigma_x^2$ and $\sigma_y^2$ are the variances, and $\sigma_{xy}$ is the covariance of the reprojected and target images. $C_1$ and $C_2$ are constants to stabilize the division. These metrics effectively capture textural conflicts and spatial misalignments induced by the lack of 3D view priors.

\textbf{Evaluation metrics for spatial intelligence and downstream utility.} Finally, to evaluate the model's capacity as a generative data engine, we benchmark the performance gain in down-stream spatial reasoning tasks. We first define Accuracy ($Acc$) as the rate of correct reasoning of Vision-Language Models:
\begin{equation}
    Acc = \frac{1}{N} \sum_{i=1}^{N} \mathbb{I} (\hat{y}_i = y_i)
\end{equation}
where $\hat{y}_i$ is the predicted answer based on 2D observations and $y_i$ is the ground truth label derived from the explicit 3D mesh and native annotations. Furthermore, we utilize Recall-Oriented Understudy for Gisting Evaluation-Longest Common Subsequence (ROUGE-L) to evaluate the structural and semantic alignment of generated textual captions. Unlike simple n-gram matching, ROUGE-L identifies the longest co-occurring sequence of tokens to account for sentence-level fluency and structural coherence:
\begin{equation}
    R_{lcs} = \frac{LCS(X, Y)}{m}
\end{equation}
\begin{equation}
    P_{lcs} = \frac{LCS(X, Y)}{n}
\end{equation}
\begin{equation}
    F_{lcs} = \frac{(1 + \beta^2) R_{lcs} P_{lcs}}{R_{lcs} + \beta^2 P_{lcs}}
\end{equation}
where $X$ and $Y$ represents the reference and candidate sequences of lengths $m$ and $n$. $LCS(X, Y)$ denotes the length of the longest common subsequence between them, and $\beta$ is a weighting parameter. These metrics collectively serve as a proxy for the model's ability to enhance the spatial perception and reasoning boundaries of intelligent agents.

\section{Results}

\subsection{Overview of MetaEarth3D and data foundation}

\begin{figure}[htbp]
    \centering
    \includegraphics[width=0.9\textwidth]{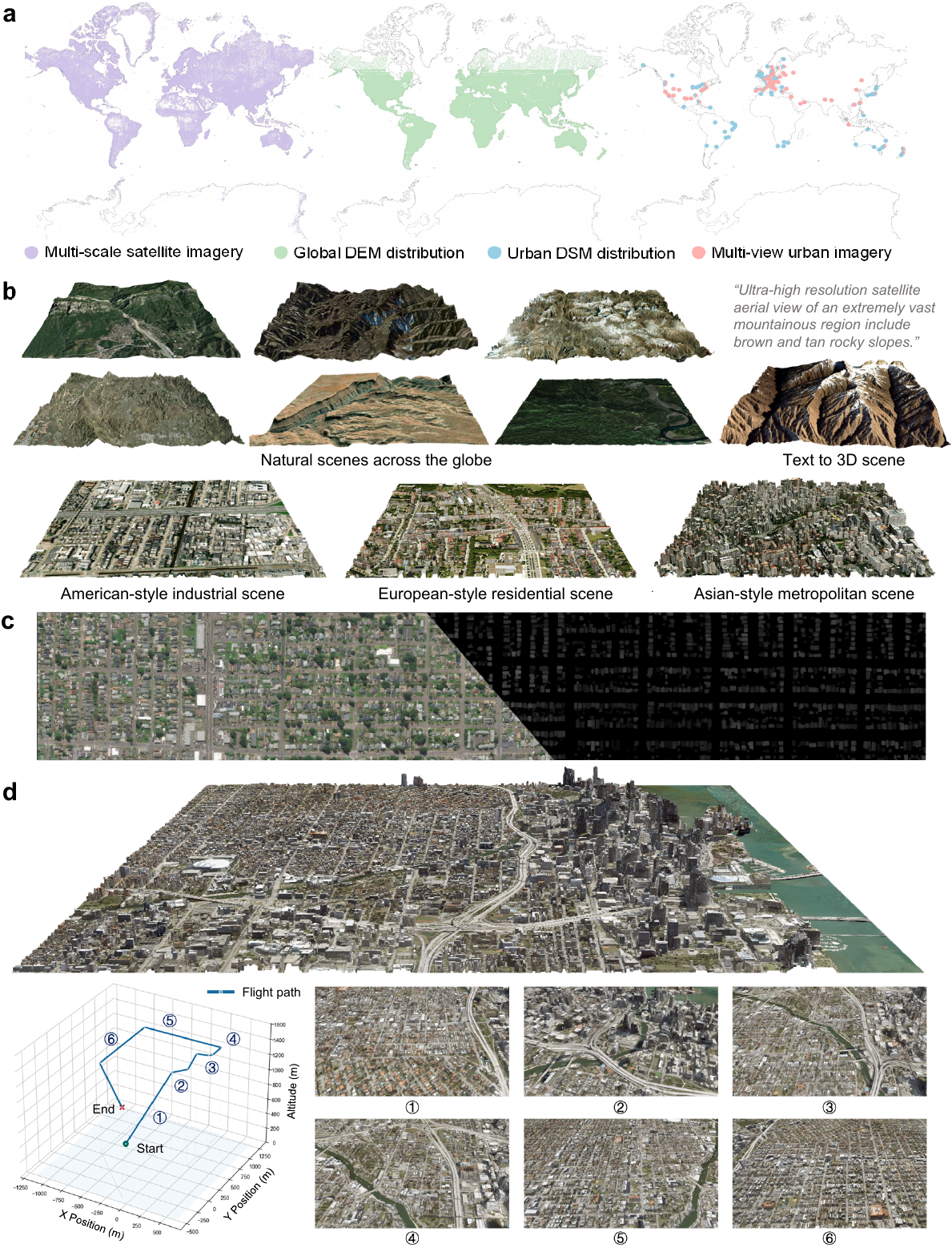}
    \caption{\textbf{Data distribution and qualitative performance of MetaEarth3D.} \textbf{a,} The global distribution of dataset supporting MetaEarth3D training and testing. \textbf{b,} Various 3D scenes generated across the globe, including mountains, deserts, plains, snow-capped mountains, and cities with distinct continental styles. All generated scenes are conditioned on either 64 m/pixel large-scale low-resolution satellite images or text descriptions. \textbf{c,} A representative sample demonstrating the pixel-wise alignment between the continuous, unbounded generated high-resolution RGB imagery (left) and its corresponding generated structural depth map (right). \textbf{d,} Unbounded exploration with view consistency. MetaEarth3D enables large-scale explicit 3D scene generation. The bottom panels display a user-defined 6-DOF flight trajectory (left) and the corresponding rendered views (1–6), demonstrating the model's ability to maintain structural and visual consistency during continuous long-range navigation.}
    \label{fig:fig2}
\end{figure}

We develop MetaEarth3D, a generative foundation model that scales 3D generation from local objects to ultra-wide Earth observation scales. With 4.6 billion parameters, the MetaEarth3D is built upon a progressive probabilistic distribution transition framework to generate multi-level, unbounded, and spatially continuous 3D scenes from either single-view satellite imagery or text descriptions. Detailed methodological descriptions are provided in the Methods section. To enable global-scale and multi-level 3D scene generation, we constructed a large-scale dataset comprising approximately 10 million images. The dataset consists of three complementary components: multi-scale satellite imagery, geo-aligned elevation maps, and multi-view urban imagery. The dataset provides both broad geographical coverage and structural diversity. Fig.~\ref{fig:fig2}a illustrates the overall data composition and spatial distribution. A summary is provided in Table~\ref{tab:dataset_overview}.

\begin{table}[htbp]
    \centering
    \caption{\textbf{Overview of our constructed dataset.} The dataset comprises three core components: multi-scale satellite imagery covering diverse global terrains, geo-aligned elevation maps for 3D geometry inference, and multi-view urban imagery for consistent texture synthesis.}
    \label{tab:dataset_overview}
    \small
    % 设置表格行高，让内容不那么拥挤
    \renewcommand{\arraystretch}{1.5}
    % \begin{tabular}{|列定义|} 使用 | 表示竖线
    \begin{tabular}{|p{1.5cm}|p{3cm}|p{1.5cm}|p{2.0cm}|p{2cm}|p{4.0cm}|}
        \hline
        \textbf{Dataset Component} & \textbf{Description} & \textbf{Spatial Resolution} & \textbf{Quantity} & \textbf{Geographic Coverage} & \textbf{Role in Training MetaEarth3D} \\ \hline
        
        Multi-scale Satellite Imagery & Optical remote sensing images annotated with geospatial metadata (lat, lon and spatial resolution) and text prompts & 256$\times$256 px patches at 64, 16, 4, 1 m/pixel & $\sim$8M filtered high-quality images (after cleaning) & Globally distributed, covering urban areas, forests, deserts, oceans, glaciers, etc. & Provides training supervision for world-scale orthographic base map synthesis and cross-scale resolution enhancement \\ \hline
        
        Geo-aligned Elevation Maps & DEM and DSM maps registered to satellite imagery & 16 m/pixel DEM + 1 m/pix DSM for urban regions & $\sim$1.2M aligned elevation maps & 79 major cities ($\sim$8,000 km$^2$) + global terrain regions & Serves as ground truth for satellite-conditioned vertical geometry inference \\ \hline
        
        Multi-view Urban Imagery & Building facades captured along circular camera trajectories & Dense angular sampling (40 viewpoints per trajectory) & $\sim$1.0M multi-view images & 116 global cities & Supervises view-conditioned lateral texture inpainting and ensures multi-view consistency \\ \hline
    \end{tabular}
\end{table}

Multi-scale satellite imagery constitutes the visual foundations of our training and sampling pipeline. We collected optical remote sensing images from available global remote sensing imagery dataset (~\cite{40, 41, 42}) at a resolution of $256 \times 256$ pixels across four spatial scales (64m, 16m, 4m, and 1m per pixel). To ensure dataset integrity and quality, we implemented a comprehensive data cleaning and annotation pipeline. First, we filtered out redundant ocean scenes and cloud-occluded images. Second, for images with degraded visual quality (e.g., noise or artifacts), we applied an image restoration model to improve their visual quality. To annotate detailed textual descriptions, we utilized GPT-4 (~\cite{14}), prompting the model with metadata including geolocation coordinates and scene classification tags to automatically synthesize high-quality captions. Concurrently, we adopted a human-in-the-loop verification strategy: we performed periodic manual spot checks on the generated captions and iteratively refined the prompt engineering based on identified errors, thereby ensuring both the efficiency and accuracy of the large-scale annotation. Following this rigorous curation process, we ultimately secured a dataset of approximately 5 million high-quality multi-resolution images. These samples encompass a broad range of global geographical environments, covering urban areas, forests, deserts, oceans, glaciers, and other representative landforms. Each image is annotated with detailed text descriptions and geographical metadata (e.g., latitude, longitude, and resolution).

Building upon the optical imagery, we further collected 1.2 million elevation maps that are spatially registered to the optical images. The elevation data comprises 16m/pixel Digital Elevation Models (DEM) sourced from the Copernicus DEM (~\cite{43}), and fine-grained 1m/pixel Digital Surface Models (DSM) covering approximately 8,000 km$^2$ across 79 major cities (derived from OpenStreetMap (~\cite{HAOZHE:link/OpenStreetMap})). Specifically, we aligned the elevation maps with the satellite images using pixel-level geospatial coordinates (latitude and longitude), resizing them to match the spatial resolution of the satellite imagery. For the elevation data itself, raw floating-point values were processed to filter invalid negative artifacts, and the valid elevation range was dynamically normalized to an 8-bit integer space to ensure stable training for the diffusion model. Finally, since satellite imagery and elevation maps inherently lack information on vertical surfaces (e.g., building facades), we additionally supplemented a multi-view urban image set from 116 cities worldwide. For each city, we set up circular camera trajectories in Google Earth Studio (~\cite{HAOZHE:link/GoogleEarth}) to capture facade views from different angles. Each trajectory consists of 40 views, yielding a total of approximately 1.0 million multi-view images that provide rich visual details for vertical surfaces.

\subsection{MetaEarth3D demonstrates high-fidelity 3D generation across global landscapes}

\textbf{Generation Diversity at Global Scale.} The MetaEarth3D demonstrates comprehensive capabilities in generating diverse 3D scenes across the globe. Benefiting from our proposed generation framework, MetaEarth3D effectively represents the diverse patterns of global scenes. Conditioned on flexible multi-modal inputs (text prompts or satellite imagery), MetaEarth3D generates realistic scenes with distinct regional characteristics, including mountains, deserts, snowfields, coastlines, and urban areas. Fig.~\ref{fig:fig1}b and Fig.~\ref{fig:fig2}b showcase representative results (additional results are provided in the Supplementary Materials). Furthermore, within the specific domain of urban generation, MetaEarth3D captures unique landscapes of different regions. As illustrated in Fig.~\ref{fig:fig2}b, using large-scale low-resolution (64 m/pixel) satellite imagery from American, European, and Asian regions as conditional base maps, MetaEarth3D effectively generates 3D urban scenes reflecting distinct regional styles (e.g., building layout and density).

\begin{figure}[htbp]
    \centering
    \includegraphics[width=0.82\textwidth]{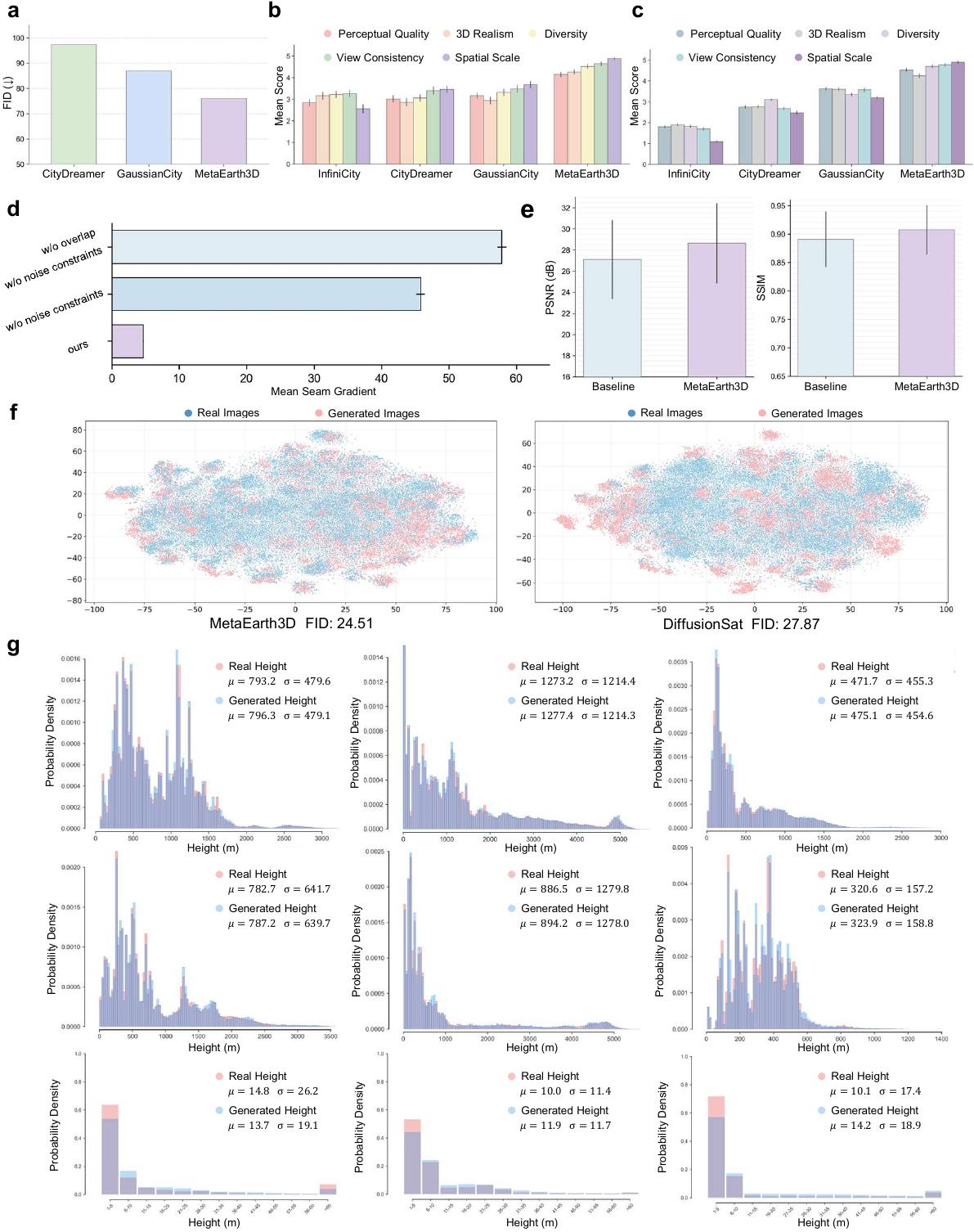}
    \caption{\textbf{Quantitative evaluation and comparison with previous methods.} \textbf{a,} Comparison of Fr\'{e}chet Inception Distance (FID) scores for generated 3D scenes with previous state-of-the-art methods. Lower FID values indicate higher generation quality. MetaEarth3D achieves the lowest FID, indicating superior visual quality compared with previous methods. \textbf{b,} Human expert evaluation results. Domain experts assessed generation quality across five dimensions: including perceptual quality, 3D realism, diversity, view consistency and spatial scale. All scores are in the range of 5, with 5 indicating the best. \textbf{c,} Multimodal Large Language Model (MLLM) evaluation results. \textbf{d,} Ablation study on scene continuity. The Mean Seam Gradient (MSG) metric demonstrates the effectiveness of the proposed unbounded generation algorithm; MetaEarth3D significantly minimizes seam artifacts compared with ablated baselines (w/o overlap or noise constraints). \textbf{e,} Assessment of multi-view texture consistency. Quantitative comparison using PSNR and SSIM, showing improvements over the baseline. \textbf{f,} t-SNE visualization of semantic feature distributions. The plots illustrate the manifold alignment between real (blue) and generated (red) images. MetaEarth3D (left) exhibits a tighter overlap with the real data distribution and a lower FID compared with DiffusionSat (right). \textbf{g,} Statistical comparison of generated versus real height maps across diverse continents and scenes. Subpanels (arranged left to right, top to bottom) display terrain elevations for Africa, Asia, Europe, North America, South America, Oceania, followed by height distributions of artificial structures in Asia, Europe and North America. The high degree of overlap (gray regions) indicates the model's capability to generate statistically realistic height maps consistent with real-world geostatistics.}
    \label{fig:fig3}
\end{figure}

\textbf{Semantically Coherent Multi-level Scene Generation.} Beyond fixed-scale scene generation, MetaEarth3D enables the creation of unified scenes across multiple observational levels within a single framework. As shown in Fig.~\ref{fig:fig1}b, MetaEarth3D generates cross-scale scenes that maintain semantic coherence, effectively encompassing macro-level terrains, medium-level cities, and fine-grained street blocks. By modeling the multi-resolution generation as a Markov transition process in scale space, the MetaEarth3D ensures that generated fine-grained details remain structurally anchored to the coarse-level semantics. This formulation establishes a robust basis for multi-level semantically aligned 3D scene generation. Furthermore, since our proposed method explicitly decouples the dimensional lifting process from the scale-space generation, MetaEarth3D enables independent control over the generation spatial resolution, achieving scale-wise controllable scene generation.

\textbf{Quantitative Analysis of Generation Quality.} We validate the generative quality of MetaEarth3D by performing a quantitative comparison with previous state-of-the-art 3D city generation methods: CityDreamer and GaussianCity. To ensure a fair comparison within the generation capabilities of the baselines, we selected Manhattan and Brooklyn as the test sites, as CityDreamer (~\cite{46}) and GaussianCity (~\cite{47}) are limited to generating large-scale oblique aerial views specifically in these regions. We rendered 15,000 images from the generated 3D scenes using diverse, randomly sampled camera trajectories. The Fr{\'e}chet Inception Distance (FID) was then calculated between these rendered views and the evaluation set (~\cite{46}) (lower FID indicates higher generation quality). As reported in Fig.~\ref{fig:fig3}a, MetaEarth3D outperforms previous SOTA methods, achieving a 12.5\% reduction in FID (from 86.94 to 76.04). Unlike the baseline methods, MetaEarth3D explicitly generates the full elements of a 3D scene, including high-resolution satellite imagery, elevation geometry, and lateral textures, resulting in generated scenes with richer visual details.

\textbf{Comprehensive Evaluation via Human and Machine Intelligence.} We conducted a comprehensive evaluation involving both human experts and Multimodal Large Language Models (MLLMs) to assess the generation quality across five critical dimensions: perceptual quality, 3D realism, diversity, view consistency, and spatial scale. For the human evaluation, we requested feedback from 50 domain experts, including researchers and practitioners from diverse fields such as architectural design, remote sensing, computer vision, and computer graphics. For the MLLM assessment, we employed Gemini 3.0 Pro (~\cite{15}) as the evaluator. We utilized a blind A/B testing protocol where evaluators scored randomly sampled multi-view renderings generated by InfiniCity (~\cite{19}), CityDreamer (~\cite{46}), GaussianCity (~\cite{47}), and our MetaEarth3D on a 5-point Likert scale. The results of the human expert assessment and the MLLM evaluation are presented in Fig.~\ref{fig:fig3}b and Fig.~\ref{fig:fig3}c, respectively. MetaEarth3D demonstrates a significant performance advantage in the human study, consistently achieving mean scores between 4.1 and 4.9 across all metrics. Evaluators reached a strong consensus regarding the model's superiority in spatial scale, while specifically highlighting the enhanced generation diversity of our results, which exhibit distinct regional characteristics absent in baseline methods.

The assessment via MLLM exhibits a high degree of alignment with human judgment, validating the high perceptual fidelity of our method. Interestingly, the MLLM evaluation displays stricter discrimination towards lower-quality generations. While human experts often assigned moderate scores to baseline methods due to subjective tolerance, the MLLM imposed severe penalties on artifacts and geometric inconsistencies. This implies that the MLLM possesses a heightened sensitivity to structural flaws and visual artifacts. These findings further validate that MetaEarth3D not only surpasses existing methods in visual fidelity but does so while achieving a significantly broader generation scope. Qualitative comparison results are provided in the Supplementary Materials.

\subsection{MetaEarth3D unites unbounded generation with explicit spatial consistency}

Achieving unbounded generation while maintaining explicit spatial consistency presents a fundamental trade-off in 3D generation. Unboundedness necessitates decomposing the scene into manageable local patches to adapt to computational constraints, yet spatial consistency demands global geometry and texture coherence. By reformulating the representation for large-scale scene generation via a designed unbounded generation strategy and a lateral appearance generation network, MetaEarth3D enables the creation of unbounded, continuous, and spatially consistent 3D meshes. Fig.~\ref{fig:fig2}c and Fig.~\ref{fig:fig2}d show some generated large-scale scenes.

\textbf{Large-scale Geometric Continuity.} The foundation for generating spatially consistent 3D scenes lies in the generation of continuous 3D geometric structures, i.e., generating globally continuous elevation maps during patch-based generation. Our previous research (~\cite{48}) validated the effectiveness of unbounded generation algorithm in pixel space for creating continuous satellite imagery. In MetaEarth3D, we extend this algorithm to the latent space of diffusion models to achieve continuous elevation map generation. Fig.~\ref{fig:fig2}c displays a generated high-resolution large-scale satellite image and its corresponding generated DSM. Visually, despite the patch-based generation strategy, both the satellite imagery and DSM maintain global continuity. We further quantitatively evaluate the algorithm's effectiveness by comparing the mean seam gradient (MSG) metrics at the stitching boundaries of DSMs generated w/ and w/o the latent space unbounded generation algorithm; lower gradient values indicate stronger continuity. As illustrated in Fig.~\ref{fig:fig3}d, the results demonstrate that the full method, i.e., employing a sliding window strategy in pixel space and constrained noise sampling in latent space, yields superior continuity at the seams, effectively enhancing the consistency of the generated height maps.

\textbf{Multi-view Texture Consistency.} We further evaluate the spatial consistency of the vertical surface textures generated by MetaEarth3D. Within the texture generation module, we introduce explicit camera pose injection and cross-view local attention, aiming to enforce visual coherence across multiple views from the latent representation. To quantitatively validate the effectiveness of our proposed method, we conducted an ablation study by establishing a baseline model that excludes both the explicit pose injection and cross-view attention modules. We employed a texture re-rendering method to measure the multi-view consistency of images generated by both MetaEarth3D and the baseline. Specifically, we back-project images generated from 8 equidistant viewpoints onto the 3D mesh surface and then re-render them from the original camera poses. We compare the re-rendered images with the originally generated ones since the differences between the re-rendered and original images' texture appear as conflicts in overlapping regions. We use Peak Signal-to-Noise Ratio (PSNR) and Structural Similarity (SSIM) to measure these pixel-level differences. As shown in Fig.~\ref{fig:fig3}e, since the baseline model lacks 3D view priors and explicit multi-view correlation modeling, it often struggles to maintain spatial consistency, yielding a PSNR of 27.11 dB and SSIM of 0.8909. In contrast, MetaEarth3D significantly reduces multi-view texture conflicts, boosting PSNR to 28.64 dB (+1.53 dB) and SSIM to 0.9076 (+0.0167). Experimental results demonstrate that our method ensures multi-view texture consistency, laying the foundation for high-fidelity 3D mesh synthesis.

\textbf{Unlocking Infinite Exploration via Explicit 3D Representations.} The explicit mesh representation offers a decisive advantage regarding spatial consistency. MetaEarth3D generates a physically explicit 3D scene, unlocking the capability for infinite, user-defined navigation within it. As shown in Fig.~\ref{fig:fig2}d, we deploy a virtual camera in a generated large-scale urban scene to execute a complex, multi-stage flight trajectory involving significant altitude changes and non-linear looping paths. Throughout this long-range navigation, the features and relative positions of generated buildings and terrain remain continuous across different viewpoints. By offering the stable environment required for continuous interaction, MetaEarth3D holds promise for downstream Earth observation and ultra-wide spatial intelligence tasks.

\subsection{MetaEarth3D mirrors the statistical laws of physical geospatial environments}

As a geospatial generative foundation model for ultra-wide Earth observation and simulation, MetaEarth goes beyond visual realism to faithfully mirror the intrinsic statistical laws of the physical world. We validated the statistical realism of the generated results by performing a statistical comparison with physical ground truth.

\textbf{Statistical Realism of Generated Satellite Imagery.} Within the MetaEarth3D pipeline, the generated satellite imagery acts as the semantic anchor for 3D scene generation, controlling the spatial layout and surface cover. Consequently, the statistical consistency of this imagery is a prerequisite for valid 3D modeling. We first evaluated the alignment between generated and real imagery (~\cite{42}) at the semantic feature level. Specifically, we utilized a pre-trained remote sensing scene classification model to extract high-dimensional semantic features from both domains and visualized their distributions via t-SNE. To complement the visual analysis with a quantitative measure, we further calculated the Fr{\'e}chet Inception Distance (FID) to measure distributional similarity (where a lower FID indicates smaller divergence). Fig.~\ref{fig:fig3}f presents the feature distributions of text-conditioned images generated by MetaEarth3D and DiffusionSat (~\cite{49}) relative to real imagery. The results show that MetaEarth3D achieves a better FID score of 24.51, and its semantic features closely overlap with those of the real images, exhibiting a well-mixed manifold distribution. This indicates that MetaEarth3D effectively captures the diverse semantic distribution of real-world scenes. In contrast, DiffusionSat displays a noticeably separated distribution and a worse FID score of 27.87.

\textbf{Statistical Consistency of Elevation Distribution.} A valid geospatial foundation model must capture the intrinsic correlation between semantic categories and physical geometry, ensuring that specific scenes exhibit their distinctive topographical or functional height distributions. To evaluate the statistical realism of MetaEarth3D, we compared the histograms of generated and real elevation data across a worldwide test set. This test set evaluation covers large-scale terrain DEMs across six continents (excluding Antarctica) and building height distributions from the Americas, Europe, and Asia. As shown in Fig.~\ref{fig:fig3}g, the generated elevation distributions closely align with the real distributions across diverse global locations, including complex non-Gaussian profiles, multi-modal distributions, and distributions with long-tailed characteristics. The consistency demonstrates that MetaEarth3D does not hallucinate geometry, but inference elevation based on semantic properties, statistically mirroring reality across global scenes.

\subsection{MetaEarth3D as a generative data engine for spatial intelligence}

Beyond visual generation, MetaEarth3D holds the potential to serve as a generative data engine for ultra-wide spatial intelligence, offering unbounded and photorealistic simulation environments. Our analysis reveals that scenes generated by MetaEarth3D possess three distinct advantages: firstly, the model generates unbounded, diverse environments ranging from dense urban layouts to complex natural terrains; secondly, it produces explicit 3D meshes, ensuring absolute multi-view consistency; thirdly, the generated scenes inherently carry self-labeled native annotations, which can be derived from generation process or explicit 3D mesh, such as precise elevation data and structured 3D relations. These attributes make MetaEarth3D fully prepared to serve as a comprehensive generative data engine, facilitating the training and verification of critical capabilities for aerospace agents for ultra-wide spatial intelligence.

\begin{figure}[htbp]
    \centering
    \includegraphics[width=\textwidth]{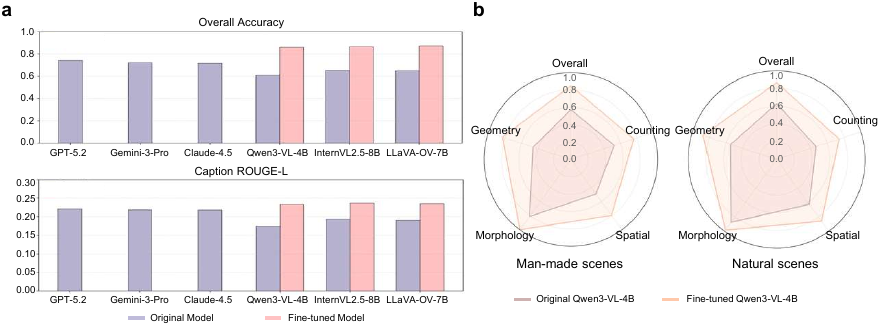}
    \caption{\textbf{MetaEarth3D as a generative data engine for spatial intelligence.} \textbf{a,} Fine-tuning efficacy across different model architectures. Quantitative comparison of spatial reasoning performance. The charts highlight the consistent improvements observed in open-source models after fine-tuning on MetaEarth3D compared with their original versions, with proprietary closed-source models included for reference. \textbf{b,} Multidimensional analysis of Qwen3-VL-4B. Radar charts illustrate the significant improvements of the fine-tuned Qwen3-VL-4B over the baseline across five dimensions in both man-made and natural scenes.}
    \label{fig:fig4}
\end{figure}

Spatial visual perception and reasoning constitute the foundational capabilities for wide-area spatial intelligence. As the cornerstone of spatial intelligence, perception forms the basis for complex interactions. In this work, to quantitatively validate the utility of MetaEarth3D as a generative data engine, we selected spatial visual perception and reasoning as representative evaluation tasks, including spatial, morphology, counting, geometry and captioning, spanning five distinct facets of 3D scene cognition, each necessitating geometric reasoning beyond 2D appearance: (1) spatial, probing 3D layout and relative placement (e.g., determining which region is higher or closer); (2) morphology, analyzing topographic forms and structural shapes defined by elevation (e.g., distinguishing ridge-valley patterns or slope gradients); (3) counting, enumerating distinct entities across varying depths and occlusions; (4) geometry, comparing explicit metric relationships such as relative height and depth ordering; and (5) captioning, generating descriptions that integrate terrain structure with land-use context. These tasks challenge an intelligent agent or vision-language models (VLMs) to perform reasoning and derive accurate answers based solely on visual observations of 3D scenes. 

However, research in ultra-wide 3D visual perception and spatial reasoning has been severely constrained by the scarcity of trustworthy 3D ground truth and the prohibitive costs of data collection and annotation. To bridge this gap, we established a human-verified automated pipeline utilizing MetaEarth3D as the data engine for synthesizing 3D scenes and ground-truth labels. We ultimately constructed a comprehensive wide-area spatial reasoning dataset comprising 7,690 samples across 60 diverse scenes, spanning natural topographies (e.g., canyons, snow-capped mountains, hills, and volcanoes) and man-made environments (e.g., residential districts, industrial zones, and rural villages). Detailed procedures for dataset construction and examples are provided in the Supplementary Materials.

We further utilized the generated samples to fine-tune open-source VLMs and subsequently evaluated their performance on the real-world UAV test set. This test set is composed entirely of real-world UAV imagery, integrating proprietary data acquired through field collection and samples collected from public web sources. At inference time, the model relies solely on 2D RGB observations as input, without access to any auxiliary 3D spatial information. Experimental results on the real-world test set demonstrate that training on MetaEarth3D data enhances the model's 3D spatial understanding. As shown in Fig. 4a, the fine-tuned open-source models achieved a significant boost in ultra-wide spatial reasoning, consistently outperforming current state-of-the-art closed-source models. The radar charts in Fig. 4b further illustrate the performance gains of Qwen3-VL-4B (~\cite{37}) across different tasks. The model exhibits a comprehensive expansion in capabilities across both natural and man-made environments, with gains being particularly pronounced in tasks necessitating explicit spatial awareness. Specifically, in geometry reasoning tasks, the model's accuracy improved by approximately +34\% (from 0.519 to 0.861). Similarly, counting accuracy improved drastically (from 0.487 to 0.749), indicating that the model learned to distinguish individual entities under occlusion using the 3D cues. Collectively, these findings indicate that our MetaEarth3D effectively extends the 3D spatial perception boundaries of current VLMs. This successful sim-to-real transfer not only underscores the quality of the generated supervision but also validates the potential of MetaEarth3D as a scalable generative data engine for physical-world spatial intelligence. 

\section{Discussion}

\textbf{Scaling Generative Foundation Models along the Spatial Axis.} MetaEarth3D pushes the boundaries of generative foundation models along the critical axis of spatial scale, significantly expanding their capability from generating bounded, object-centric assets to unbounded, world-scale 3D environments. Our experiments show that despite the immense diversity of the Earth's surface, these complex visual patterns can be effectively compressed into a unified neural representation while maintaining semantic consistency across scales. This suggests that the model goes beyond simple memorization; instead, it appears to learn the fractal nature of geospatial data, generalizing texture rules across different orders of magnitude. Moreover, by generatively modeling the distribution of visual textures and geometry rather than relying on one-to-one mappings, the model appears to capture the ‘grammar’ of the physical world, reflecting the natural dependencies between surface appearance and 3D geometry.

\textbf{Towards Simulation-Ready World Models for Spatial Intelligence.} Recent advancements in generative foundation models have driven the emergence of the concept of a world simulator or world model. A central debate persists regarding the optimal path for these models: whether to rely on implicit representations, exemplified by video generation models, or explicit 3D representations. Implicit methods, such as recent large-scale video generators, excel at modeling the temporal dynamics of pixels to produce visually high-fidelity sequences. However, they often struggle to maintain long-range spatial consistency and lack the explicit structural support required for physical interaction and reasoning. In contrast, world models based on explicit 3D representations generate not merely visual content but also intrinsic spatial data, including depth and geometry, ensuring that the synthesized world is grounded in a stable coordinate system. In this context, MetaEarth3D can be viewed as a realization of such an explicit world model. Beyond generating photorealistic environments, it inherently synthesizes spatially aligned annotations, such as geometric elevation maps and surface normals, as a direct byproduct of its generative process. While this work has primarily demonstrated the model’s effectiveness in visual perception tasks, its true potential lies in its ability to provide a simulation-ready environment. By integrating with physics-based simulation engines, MetaEarth3D holds the potential to open new avenues for broader spatial intelligence tasks, including autonomous navigation, path planning, and robotic control. We envision MetaEarth3D serving as a scalable, interactive training ground, enabling aerospace agents to develop and generalize wide-area spatial intelligence across vast, diverse, and physically realistic unbounded environments.

\textbf{Ethical Implications and Secure Data Synthesis.} Regarding data security and privacy, MetaEarth3D offers a promising solution to the dilemma of data scarcity versus sensitivity in geospatial analytics. High-resolution geospatial data is often restricted due to security concerns or privacy regulations, limiting the training resources available for public research. By synthesizing "non-existent" yet physically plausible environments, our model provides a privacy-preserving alternative: it enables the training of robust aerospace agents without exposing sensitive real-world geographical information. While the potential for misuse, such as the creation of misleading geospatial information, necessitates future research into provenance tracking and 3D watermarking, the current framework suggests that synthetic data holds the promise of serving as a secure, unclassified surrogate for developing wide-area spatial intelligence.

\textbf{Towards High-Dimensional Synthesis: The MetaEarth-XD Prospect.} While MetaEarth3D demonstrates the potential of extending generative foundation models to ultra-wide spatial extents, it currently remains confined to static, single-modal (optical) environments. However, the physical reality is inherently high-dimensional: it is not only temporally evolving but also rich in multi-spectral and multi-source information beyond the visible spectrum. A primary limitation of our current model is the absence of these additional dimensions, which restricts the simulation of both dynamic phenomena (such as weather variability and seasonal cycles) and complex sensor modalities. To address this, we envision the next evolution of our framework, MetaEarth-XD, which aims to scale the generative foundation model along multiple critical axes. This advancement involves expanding into the temporal dimension to capture dynamic changes, while simultaneously extending into spectral and modal dimensions to synthesize infrared, hyperspectral, and Synthetic Aperture Radar data. Such a multidimensional expansion would transform the current model from a static spatial generator into a comprehensive high-dimensional world simulator.

\newpage
\section*{Data availability}
The study utilized several publicly available datasets to train and validate multi-scale satellite imagery generation models, including Git-10M (available at \url{https://huggingface.co/datasets/lcybuaa/Git-10M}), Million-AID (available at \url{https://captain-whu.github.io/DiRS/}), and NWPU-Captions (available at \url{https://github.com/HaiyanHuang98/NWPU-Captions}). Global DEM data were acquired from the elevation product Copernicus DEM GLO 30 (available at \url{https://portal.opentopography.org/datasetMetadata?otCollectionID=OT.032021.4326.1}). DSM data were obtained from OpenStreetMap (available at \url{https://www.openstreetmap.org/}). Multi-view oblique imagery was collected using Google Earth Studio (available at \url{https://www.google.com/earth/studio/}). For the benchmark of urban scene generation fidelity, the CityDreamer dataset was selected (available at \url{https://gateway.infinitescript.com/s/GoogleEarth}).

\section*{Code availability}
The source code is available from the corresponding author upon reasonable request.

\bibliographystyle{unsrtnat}
\bibliography{references}  %%% Uncomment this line and comment out the ``thebibliography'' section below to use the external .bib file (using bibtex) .




% 	\bibitem{keshet2016prediction}
% 	Keshet, Renato, Alina Maor, and George Kour.
% 	\newblock Prediction-Based, Prioritized Market-Share Insight Extraction.
% 	\newblock In {\em Advanced Data Mining and Applications (ADMA), 2016 12th International 
%                       Conference of}, pages 81--94,2016.

% \end{thebibliography}

\newpage

\begin{center}
    {\huge \textbf{MetaEarth3D: Unlocking World-scale 3D Generation with Spatially Scalable Generative Modeling}}
    
    \vspace{0.3cm}
    
    {\Large Supplementary Material}
    
\end{center}

\vspace{0.2cm}

\appendix

\section{Fig. 1a References}

% --- 强制修改编号前缀为 "Supplementary Table " ---
\setcounter{table}{0}
\renewcommand{\thetable}{Supplementary Table \arabic{table}}
% 如果你还想去掉默认的 "Table" 字样重复（防止变成 Table Supplementary Table 1）
% 建议使用 caption 宏包提供的格式化功能，或者简单地这样写：
\renewcommand{\tablename}{Supplementary Table}
\renewcommand{\thetable}{\arabic{table}}

% --- 插入表格 ---
\begin{table}[ht]
\centering
\caption{Summary of Figure.1a References} % 表格标题
\label{tab:references}
\small % 减小字号以适应大量文本
\begin{tabular}{|l|p{13cm}|} % 第一列左对齐，第二列固定宽度并自动换行
\hline
\textbf{Method} & \textbf{Reference} \\ \hline
GAN & Goodfellow, I. et al. Generative adversarial networks. \textit{Commun. ACM} \textbf{63}, 139–144 (2020). \\ \hline
ProGAN & Gao, H., Pei, J. \& Huang, H. Progan: Network embedding via proximity generative adversarial network. In \textit{Proc. 25th ACM SIGKDD Int. Conf. on Knowledge Discovery \& Data Mining (KDD)} 1308-1316 (ACM, 2019). \\ \hline
BigGAN & Brock, A., Donahue, J. \& Simonyan, K. Large scale GAN training for high fidelity natural image synthesis. In \textit{Proc. 7th International Conference on Learning Representations (ICLR)} (2019). \\ \hline
StyleGAN2 & Karras, T. et al. Analyzing and improving the image quality of StyleGAN. In \textit{Proc. IEEE/CVF Conf. on Computer Vision and Pattern Recognition (CVPR)} 8110-8119 (IEEE, 2020). \\ \hline
DALL-E 1 & Ramesh, A. et al. Zero-shot text-to-image generation. In \textit{Proc. 38th Int. Conf. on Machine Learning} 8821-8831 (PMLR, 2021). \\ \hline
Stable Diff 1.5 & Rombach, R. et al. High-resolution image synthesis with latent diffusion models. In \textit{Proc. IEEE/CVF Conf. on Computer Vision and Pattern Recognition (CVPR)} 10684-10695 (IEEE, 2022). \\ \hline
SDXL & Podell, D. et al. Sdxl: Improving latent diffusion models for high-resolution image synthesis. Preprint at \url{https://arxiv.org/abs/2307.01952} (2023). \\ \hline
Stable Diffusion 3 (SD3) & Esser, P. et al. Scaling rectified flow transformers for high-resolution image synthesis. In \textit{Proc. 41st Int. Conf. on Machine Learning (PMLR, 2024)}. \\ \hline
FLUX.1 & Batifol, S. et al. FLUX.1 Kontext: Flow Matching for In-Context Image Generation and Editing in Latent Space. Preprint at \url{https://arxiv.org/abs/2506.15742} (2025). \\ \hline
DreamFusion & Poole, B., Jain, A., Barron, J. T. \& Mildenhall, B. Dreamfusion: Text-to-3d using 2d diffusion. Preprint at \url{https://arxiv.org/abs/2209.14988} (2022). \\ \hline
Magic3D & Lin, C. H. et al. Magic3d: High-resolution text-to-3d content creation. In \textit{Proc. IEEE/CVF Conf. on Computer Vision and Pattern Recognition (CVPR)} 300-309 (IEEE, 2023). \\ \hline
DreamGaussian & Tang, J., Ren, J., Zhou, H., Liu, Z. \& Zeng, G. Dreamgaussian: Generative gaussian splatting for efficient 3d content creation. In \textit{Proc. 12th International Conference on Learning Representations (2024)}. \\ \hline
Trellis & Xiang, J. et al. Structured 3d latents for scalable and versatile 3d generation. In \textit{Proc. IEEE/CVF Conf. on Computer Vision and Pattern Recognition (CVPR)} 21469-21480 (IEEE, 2025). \\ \hline
Hunyuan3D 2 & Zhao, Z. et al. Hunyuan3D 2.0: Scaling diffusion models for high resolution textured 3D assets generation. Preprint at \url{https://arxiv.org/abs/2501.12202} (2025). \\ \hline
Gen-2 & Gen-2: Generate novel videos with text, images or video clips. Runway. \url{https://runwayml.com/research/gen-2} (2023). \\ \hline
Pika 1.0 & Pika: An idea-to-video platform that brings your creativity to motion. Pika Labs. \url{https://pika.art} (2023). \\ \hline
GAIA-1 & Hu, Anthony, et al. Gaia-1: A generative world model for autonomous driving. Preprint at \url{https://arxiv.org/abs/2309.17080} (2023). \\ \hline
Sora & Brooks, T. et al. Video generation models as world simulators. OpenAI. \url{https://openai.com/index/sora/} (2024). \\ \hline
GaussianCity & Xie, H. et al. Generative Gaussian splatting for unbounded 3D city generation. In \textit{Proc. IEEE/CVF Conf. Comput. Vis. Pattern Recognit (CVPR)}. 6111-6120 (IEEE, 2025). \\ \hline
\end{tabular}
\end{table}

\section{Dataset Details}

% 1. 设置显示的标签文字 (你之前的设置)
\renewcommand{\figurename}{Supplementary Figure}

% 2. 重置编号并设置编号格式 (显示为 Sx)
\setcounter{figure}{0}
\renewcommand{\thefigure}{\arabic{figure}}

% 3. 关键：修复跳转冲突 (让 hyperref 产生唯一的锚点名)
\makeatletter
\renewcommand{\theHfigure}{Supp.\thefigure}
\makeatother

\subsection{Dataset for MetaEarth3D Training and Testing}

\begin{figure}[htbp]
    \centering
    \includegraphics[width=\textwidth]{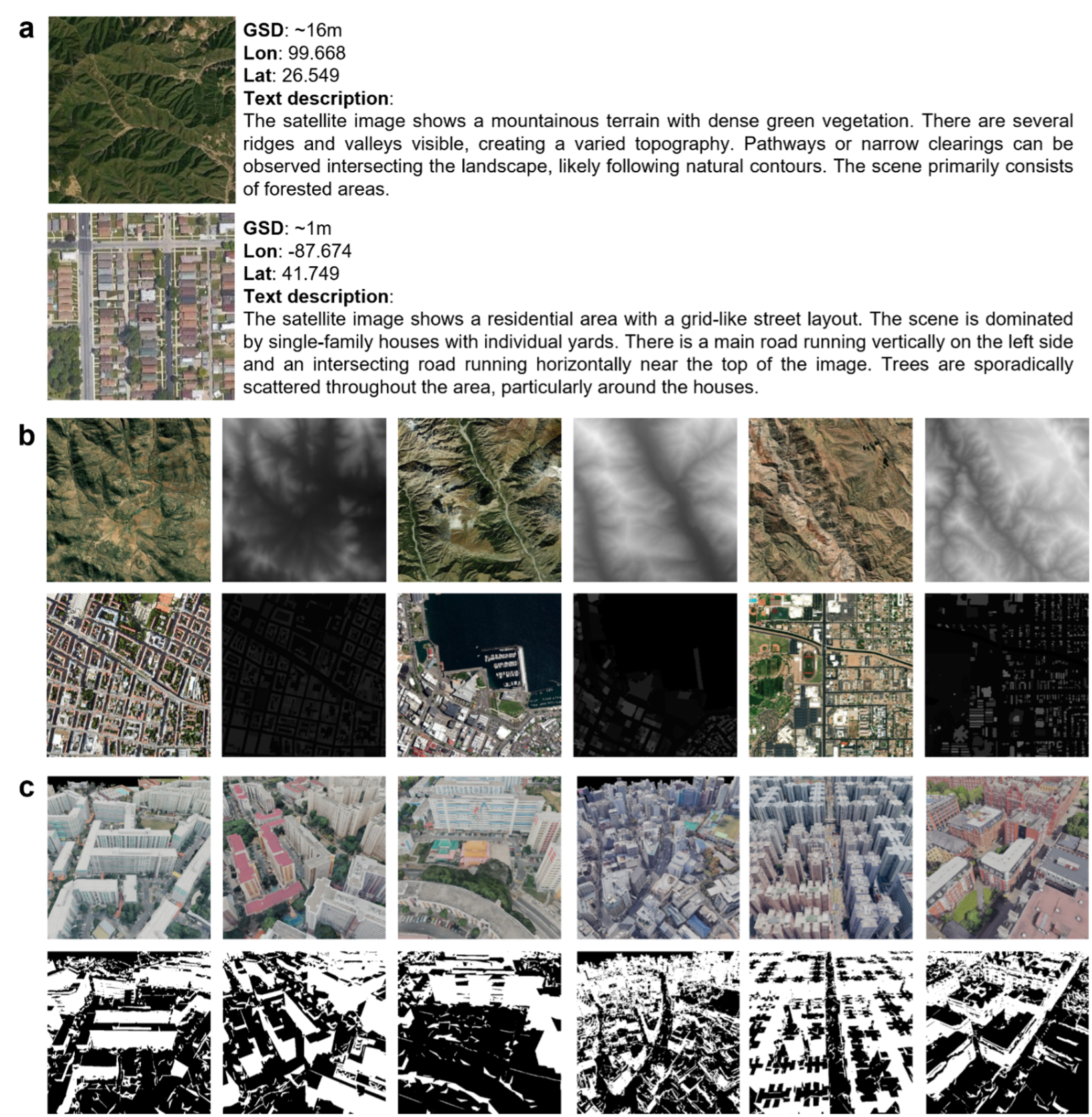} 
    \caption{\textbf{Samples of the training image data for MetaEarth3D.} \textbf{a,} Samples of 2D satellite images used for training. Each image is annotated with its spatial resolution (GSD), geographic coordinates, and a text description. \textbf{b,} Geo-aligned Elevation Maps. The top row displays paired satellite imagery and elevation maps at 16 m/pixel; the bottom row shows paired satellite imagery and height maps at 1 m/pixel. \textbf{c,} The top row presents oblique RGB images, and the bottom row displays the corresponding annotation masks for lateral textures.}
    \label{supfig1}
\end{figure}

Building upon the dataset overview in the main text, we provide a detailed statistical analysis of the 10 million training samples to demonstrate scene diversity. Representative samples are provided in Supplementary Figure \ref{supfig1}. We further analyzed the distribution of lateral texture masks within the oblique views. As visualized in the histogram in Supplementary Figure \ref{supfig2}, under consistent oblique observation angles, the proportion of lateral texture serves as a direct proxy for building density and verticality. The resulting distribution spans a wide spectrum, confirming that the dataset covers scenarios ranging from sparse areas with low texture ratios to dense, complex regions with high texture ratios. This broad coverage of scene morphologies is essential for equipping MetaEarth3D with robust generalization capabilities across heterogeneous environments.

\begin{figure}[htbp]
    \centering
    \includegraphics[width=0.8\textwidth]{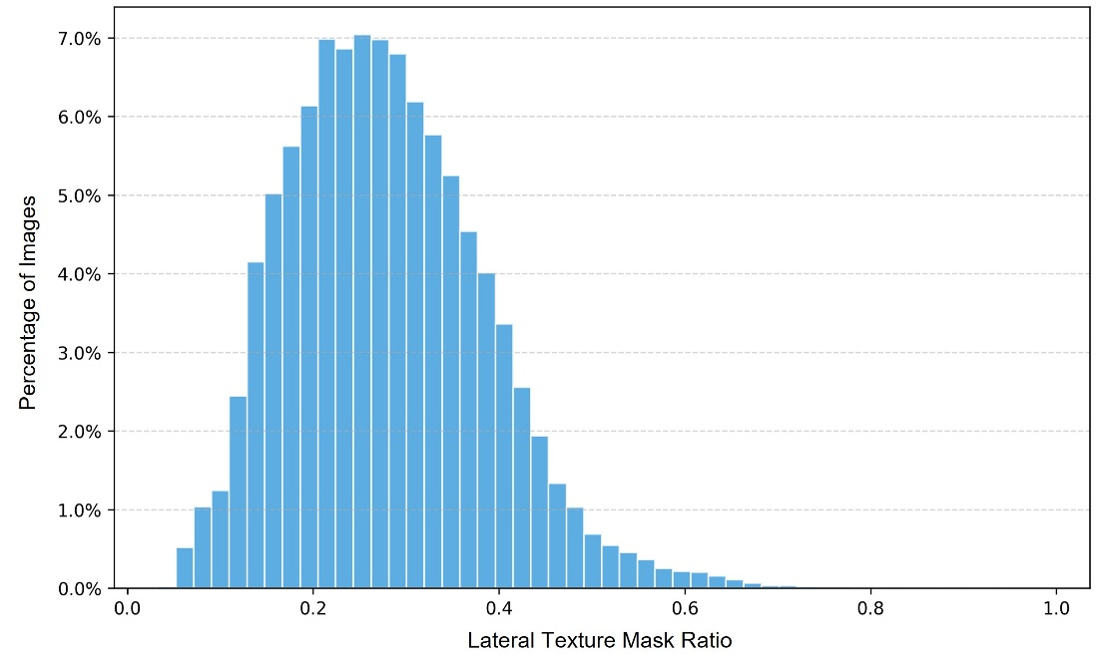} 
    \caption{\textbf{Distribution of lateral texture mask ratios.} The histogram illustrates the frequency distribution of lateral texture pixel proportions across the training dataset. This metric serves as a quantitative proxy for building density and scene complexity.}
    \label{supfig2}
\end{figure}

To comprehensively evaluate the model across different modalities, we utilized three distinct test sets: 
(i) 3D scene generation benchmark: To quantitatively compare our method with state-of-the-art baselines, we followed the evaluation settings established in CityDreamer (~\citesup{s1}). This ensures a strictly aligned benchmark for assessing 3D generation quality. 
(ii) To validate the statistical realism and semantic alignment of generated satellite imagery, we utilized the NWPU-Caption dataset (~\citesup{s2}), comprising 31,500 image-caption pairs. We performed zero-shot evaluation, using the text descriptions to generate imagery with MetaEarth3D and DiffusionSat (~\citesup{s3}) without prior fine-tuning. This setting rigorously tests the models' ability to generate semantically accurate textures from unseen prompts. 
(iii) To evaluate the statistical realism of height map generation, we constructed a test set sampled from our collected 16m/pixel DEM and 1m/pixel DSM data. Specifically, we randomly sampled approximately 6,000 images from the dataset to serve as a held-out evaluation set for quantifying the statistical realism of the generated height maps.

\subsection{Downstream Task Dataset Construction}

\begin{figure}[htbp]
    \centering
    \includegraphics[width=\textwidth]{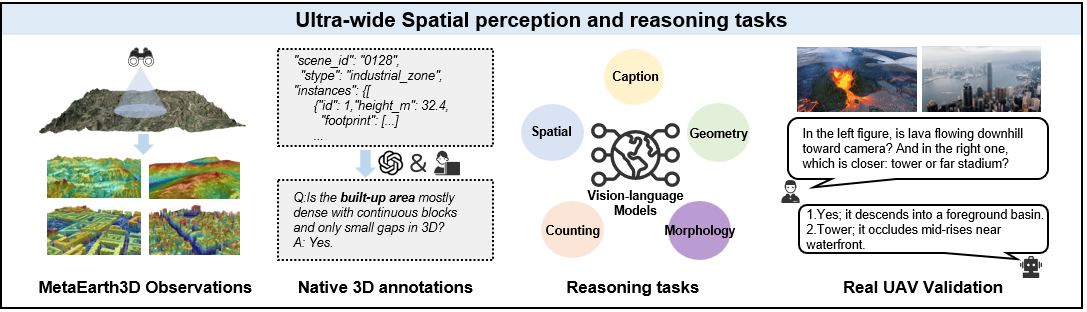} 
    \caption{\textbf{Pipeline of downstream task dataset construction.} MetaEarth3D renders calibrated UAV-style observations and provides structured 3D geometric ground truth, which is converted into 3D spatial reasoning supervision via GPT-assisted annotation with human checking. The resulting supervision supports 3D-aware adaptation of 2D VLMs across five task families (spatial, morphology, counting, geometry, caption), and is evaluated on real UAV images without mesh access at test time.}
    \label{supfig3}
\end{figure}

In the article, we employed MetaEarth3D as a generative data engine to synthesize a multi-scenario 3D visual perception and reasoning dataset, validating its efficacy in downstream spatial intelligence tasks. Supplementary Figure \ref{supfig3} illustrates the dataset construction pipeline. The dataset construction pipeline and details are as follows: We first synthesize explicit 3D meshes covering both natural and man-made environments, including natural scenes (canyon, snow mountain, hill, peak, bare rock without vegetation, and volcano) and man-made scenes (residential area, industrial area, CBD, and rural settlement). For each mesh, we simulate UAV remote-sensing viewpoints and render calibrated RGB observations along designed trajectories, producing image-pose aligned data with controlled geometric and photometric conditions. For each rendered view, the explicit mesh makes it possible to obtain geometry signals that are typically unavailable or unreliable in real UAV imagery: (i) height-related cues (scene/region height distributions, height-map visualizations), and (ii) structured 3D relations (e.g., relative height ordering, depth ordering, and cross-view consistent geometric relations). These mesh-derived cues are organized as structured 3D ground truth and are used only for annotation and training supervision.

Each RGB image is paired with exactly five questions (strictly balanced per image), one per task family: spatial (3D layout and relative placement such as higher/lower or nearer/farther), morphology (terrain/built-form patterns governed by elevation/relief, e.g., ridge-valley structures or continuous blocks), counting (numerical counting under depth/occlusion), geometry (explicit 3D relations such as relative height ordering and cross-view consistency), and caption (a concise description integrating terrain structure with human land use). This yields 7,690 question-answer (QA) pairs over 1,838 images from 60 scenes. The domain split is 53.7\% man-made and 46.3\% natural. Answers span five canonical formats: boolean, number, text, choice, and list. QA drafts are produced using GPT-5.2 (~\citesup{s4}) conditioned on (i) the RGB image and (ii) mesh-derived geometric visualizations (height map / auxiliary geometric views / overlays), followed by human verification to ensure correctness and to remove ambiguous cases.

\begin{figure}[htbp]
    \centering
    \includegraphics[width=\textwidth]{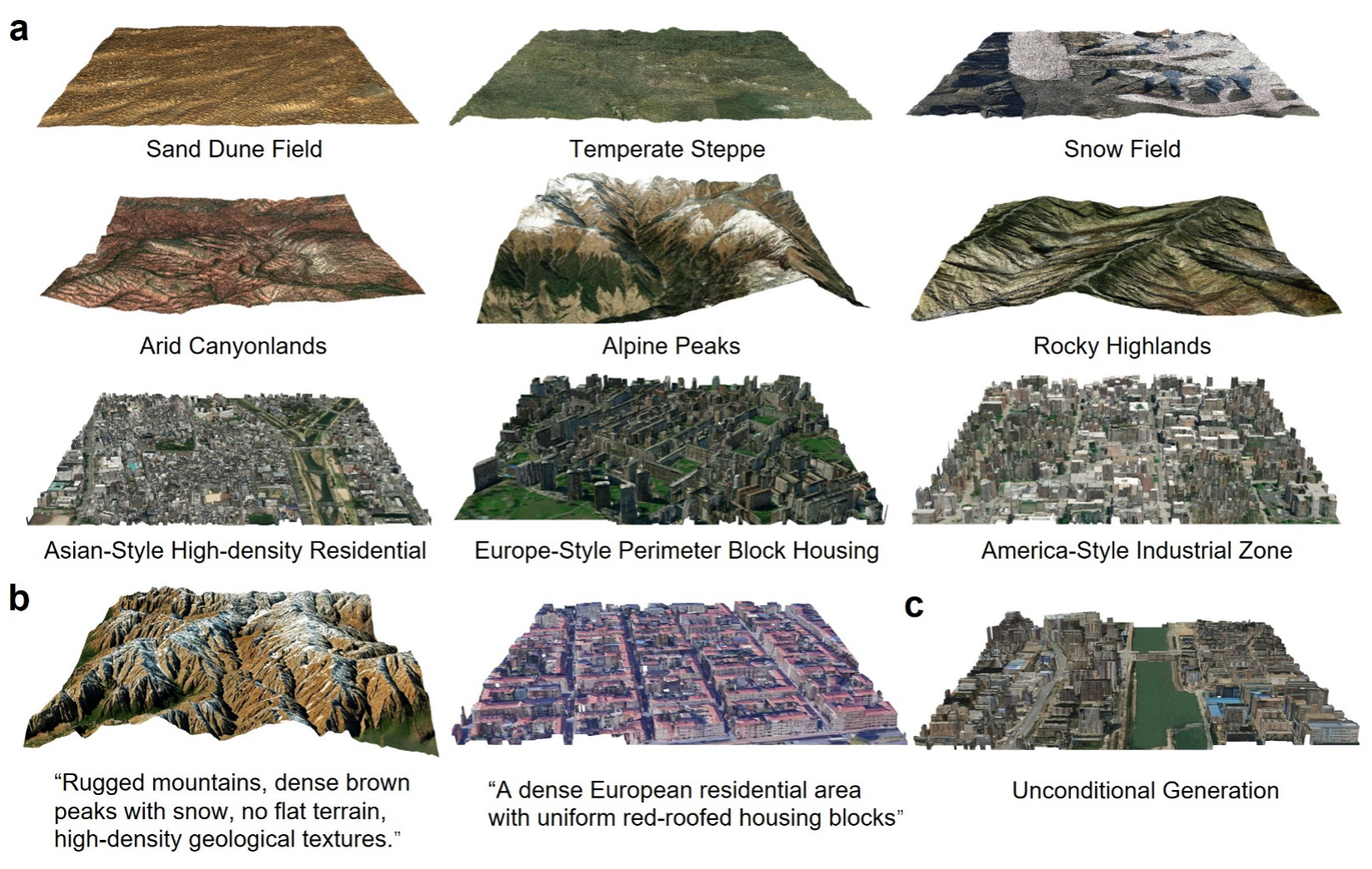} 
    \caption{\textbf{Multi-modal driven diverse 3D scene generation across the globe.} \textbf{a,} Examples of 3D scenes generated by MetaEarth3D, encompassing global natural landscapes (e.g., deserts, snowfields, and mountains) and man-made structures exhibiting distinct regional styles. All scenes are conditioned on large-scale satellite imagery with a spatial resolution of 64m/pixel. \textbf{b,} Samples of text-to-3D scene generation. \textbf{c,} Unconditional generation results, where a generic text prompt ("a satellite image") was employed during the initial satellite imagery generation stage.}
    \label{supfig4}
\end{figure}

\section{MetaEarth3D Experimental Results}

In this supplementary material, we provide additional visualizations demonstrating the diverse generation capabilities of MetaEarth3D, alongside a comprehensive qualitative comparison against previous state-of-the-art methods. 

\subsection{MetaEarth3D Scene Generation Results}

\begin{figure}[htbp]
    \centering
    \includegraphics[width=\textwidth]{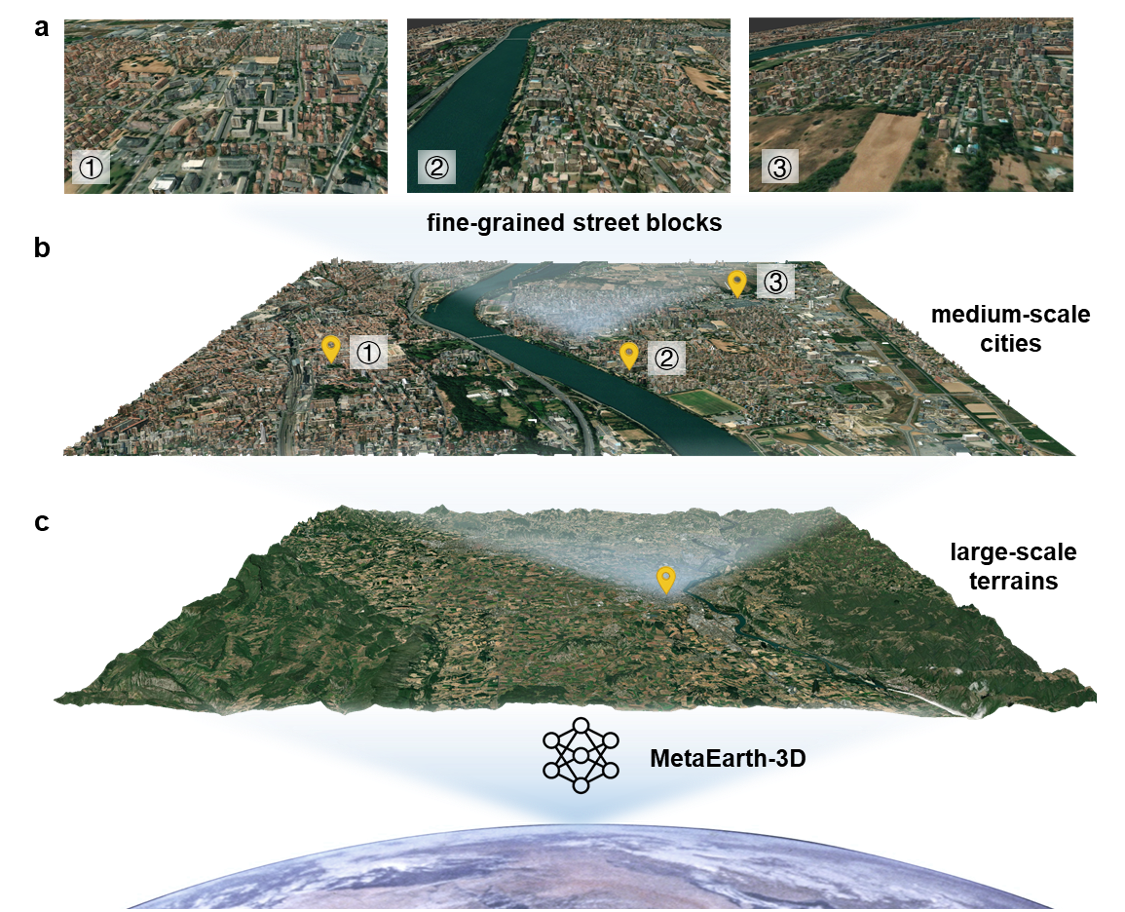} 
    \caption{\textbf{Multi-level and Unbounded 3D scene generation result. a-c,} Arranged from bottom to top, these visualizations span three distinct scales: a macro-scale terrain covering approximately 4,300 km² (c), a median-scale urban scene (b), and a fine-grained city block scene (a).}
    \label{supfig5}
\end{figure}

\textbf{Diverse scene generation across the globe.} Supplementary Figure \ref{supfig4}a illustrates a diverse array of 3D scenes generated by MetaEarth3D. All generated scenes are conditioned on large-scale satellite imagery with a spatial resolution of 64m/pixel. Supplementary Figure \ref{supfig4}b presents the results of text-to-3D scene generation. Supplementary Figure \ref{supfig4}c displays the outcomes of unconditional generation, where the generic text prompt "a satellite imagery" was employed during the initial satellite image generation stage.

\textbf{Multi-level and unbounded scene generation.} Supplementary Figure \ref{supfig5} showcases an additional set of large-scale, multi-resolution scene generations generated by MetaEarth3D. Arranged from bottom to top, these visualizations span three distinct scales: a macro-scale terrain covering approximately 4,300 km², a median-scale urban scene, and a fine-grained city block scene.

\subsection{Qualitative Comparison with Previous SOTA Methods}

\begin{figure}[htbp]
    \centering
    \includegraphics[width=\textwidth]{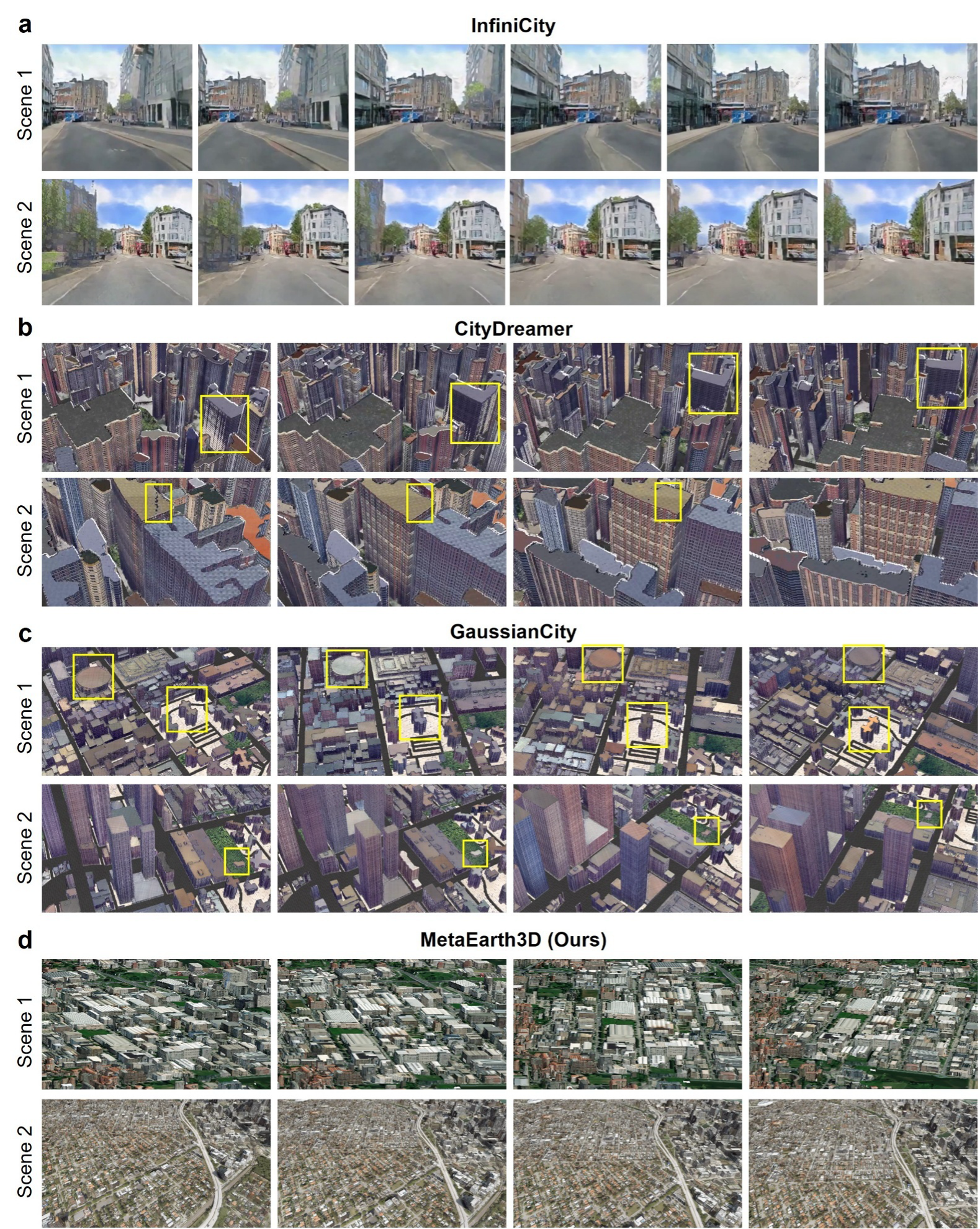} 
    \caption{\textbf{Qualitative comparison with previous generative methods. a-d,} Multi-view renderings of 3D scenes generated by InfiniCity (a), CityDreamer (b), GaussianCity (c), and MetaEarth3D (d). Yellow bounding boxes highlight regions exhibiting view-inconsistent visual textures (or content drift) across different viewpoints.}
    \label{supfig6}
\end{figure}

Supplementary Figure \ref{supfig6} provides a visual comparison of urban scenes rendered by previous methods: InfiniCity (~\citesup{s5}), CityDreamer (~\citesup{s1}), and GaussianCity (~\citesup{s6}). Specifically, InfiniCity relies on an octree-based voxel representation, CityDreamer utilizes a neural field representation, and GaussianCity employs a 3D Gaussian Splatting representation. As observed in the figure, scenes generated by InfiniCity suffer from severe geometric distortion and blurriness. While CityDreamer and GaussianCity achieve better textural clarity, they exhibit monotonous lateral textures on buildings, with roof textures being almost entirely absent. Furthermore, as these baselines lack explicit 3D representations, they suffer from noticeable content drift when the viewing angle changes (highlighted by the yellow bounding boxes). In contrast, MetaEarth3D generates scenes with rich, high-fidelity textures. Crucially, by leveraging an explicit mesh-based representation, our method ensures strict multi-view consistency.

\subsection{Qualitative results of the downstream ultra-wide spatial reasoning task}

\begin{figure}[htbp]
    \centering
    \includegraphics[width=\textwidth]{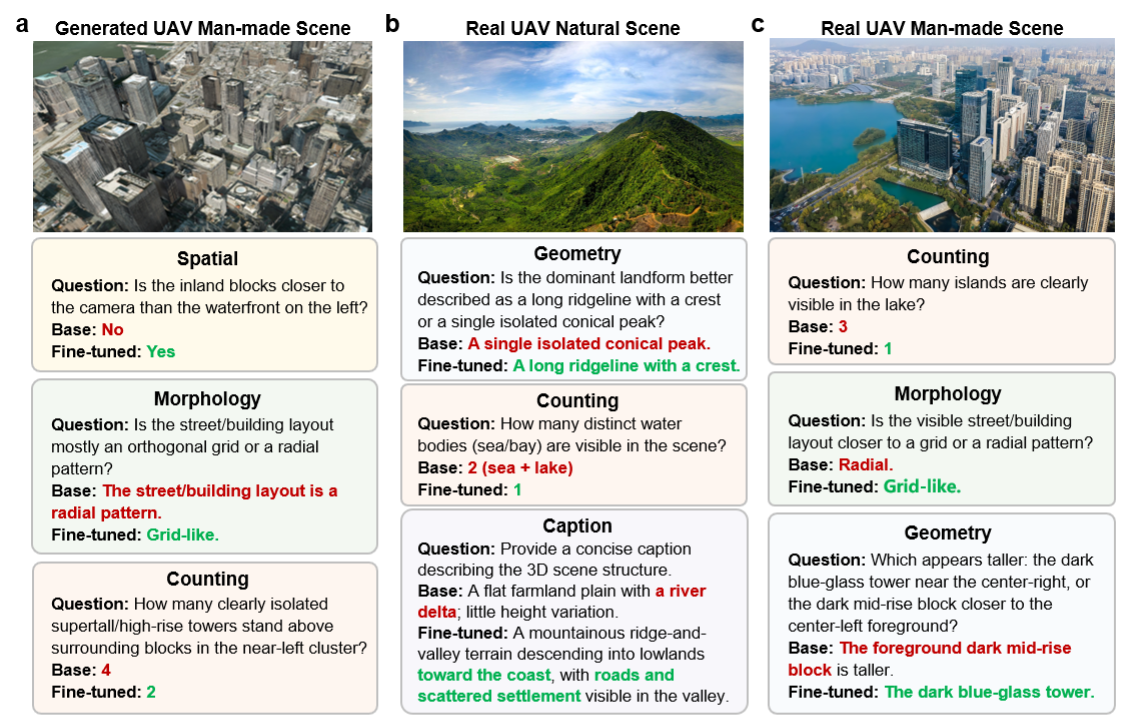} 
    \caption{\textbf{Qualitative results of the downstream ultra-wide spatial reasoning task. a-c,} Representative inference examples on a generated scene (a) and real-world UAV imagery (b, c). Green text indicates correct responses, while red text indicates incorrect ones. The comparison highlights the fine-tuned model's ability to correctly interpret complex 3D spatial relationships, geometric structures, and object counts. In contrast, the original baseline exhibits hallucinations or fails to grasp specific spatial contexts. These results demonstrate the fidelity of the synthesized supervision, indicating that spatial priors learned from MetaEarth3D facilitate generalization to real-world environments.}
    \label{supfig7}
\end{figure}

Supplementary Figure \ref{supfig7} provides representative inference examples on a generated scene (Supplementary Figure \ref{supfig7}a) and real-world UAV imagery (Supplementary Figure \ref{supfig7}b, Supplementary Figure \ref{supfig7}c). Green text indicates correct responses, while red text indicates incorrect ones. The comparison highlights the fine-tuned model's ability to correctly interpret complex 3D spatial relationships, geometric structures, and object counts. In contrast, the original baseline exhibits hallucinations or fails to grasp specific spatial contexts. These results demonstrate the fidelity of the synthesized supervision, indicating that spatial priors learned from MetaEarth3D facilitate generalization to real-world environments.

\section{Experiment settings of MetaEarth3D}

All models and experiments were implemented using Python and PyTorch. The specific model designs and implementation details are as follows:

(i) The text-to-remote-sensing image generation module we trained is implemented based on the Stable Diffusion architecture, with a total of 1.3 billion parameters. Training was conducted on a machine equipped with 8 NVIDIA A100 GPUs, with the learning rate set to $1 \times 10^{-4}$ and a batch size of 1024. During the inference phase, the number of sampling steps was set to 40, enabling the generation of $256 \times 256$ resolution images. For the resolution-guided recursive diffusion network, we designed a pixel-space U-Net-like architecture to predict noise, containing a total of approximately 1.0 billion parameters. The feature maps undergo four downsampling and upsampling stages, with channel dimensions expanding or reducing by factors of [1, 2, 4, 8] relative to the base channel number, which is set to 160. The model was trained on 4 NVIDIA A800 GPUs, with a batch size of 128 and a learning rate of $1 \times 10^{-5}$.

(ii) Our proposed geometry generator shares a similar architecture with InstructCV (~\citesup{s7}), with a total of 1.0 billion parameters. A pre-trained CLIP model was employed to encode the task prompt. At each diffusion timestep, the height map latent code is concatenated along the channel dimension with the conditional satellite imagery encoded by a pre-trained VAE, and then input into the denoising network. The model was trained on 8 NVIDIA RTX 4090 GPUs. The batch size was set to 128, using the AdamW optimizer with a learning rate of $1 \times 10^{-5}$.

(iii) Our proposed texture generator contains 1.3 billion parameters and is structurally similar to AnimateDiff (~\citesup{s8}). Specifically, the input convolutional layer was expanded to 9 channels to accommodate the joint input of the noisy latent code (4 channels), the encoded masked multi-view condition images (4 channels), and the lateral texture mask (1 channel). We replaced the attention modules in the original 3D Transformer with our proposed cross-view attention; this process introduces no additional parameters. The model was trained on 4 NVIDIA A800 GPUs. The total batch size was set to 16, using the AdamW optimizer with a learning rate of $7 \times 10^{-5}$.

\section{Experiment settings of the Downstream Task}

We adapt open-source 2D VLM backbones (~\citesup{s9, s10, s11}) via LoRA-based supervised fine-tuning (~\citesup{s12}) using the same Qwen-style single-turn message format as training input. The implementation follows our finalized training script settings.

We fine-tune each open-source backbone with LoRA-based SFT using a single-turn Qwen-style multimodal format. The LoRA adapters use r=16, $\alpha$=32, dropout=0.05, and no bias terms, and are injected into the standard projection layers of the language backbone. Training is conducted in fp16 with a maximum sequence length of 2048 tokens. We use a per-device batch size of 1 with 16-step gradient accumulation, learning rate $5 \times 10^{-5}$, warmup ratio 0.05, weight decay $1 \times 10^{-5}$, and 5 training epochs. Multimodal inputs are kept intact by disabling column pruning. The data collator constructs chat-formatted prompts, loads the paired RGB image, and forms the language-model loss by masking padding tokens in the label sequence.

For downstream validation, models are evaluated on a real test set using RGB images only at inference time. We report overall reasoning accuracy, per-task accuracy for spatial/morphology/counting/geometry, and caption quality measured by ROUGE-L. Closed-source models are included as direct-inference references under the same protocol, while open-source backbones are compared before and after LoRA adaptation to quantify the transfer of mesh-derived 3D supervision under synthetic-to-real domain shift.

\section{Supplementary movies}

\subsection{Supplementary movie 1. Multi-modal driven world-scale 3D scene generation}

This video demonstrates the capability of MetaEarth3D to generate diverse 3D scenes across the globe, driven by distinct input modalities (e.g., text prompts and satellite imagery).

\subsection{Supplementary movie 2. Multi-level and large-scale 3D scene generation}

In this video, we demonstrate the capability of MetaEarth3D to generate large-scale environments across multiple levels of detail. We specifically highlight the semantic consistency maintained throughout the multi-level generation process.

\subsection{Supplementary movie 3. Interactive exploration of unbounded explicit 3D scenes}

In this video, we demonstrate MetaEarth3D's capability to generate unbounded, explicit 3D scenes, enabling interactive, user-defined, and continuous free-viewing.

\bibliographystylesup{unsrtnat}
\bibliographysup{references}

\end{document}